\documentclass[10pt,journal,compsoc]{IEEEtran}

\ifCLASSOPTIONcompsoc
  \usepackage[nocompress]{cite}
\else
  \usepackage{cite}
\fi
\ifCLASSINFOpdf
\else
\fi
\usepackage{cite}
\usepackage{amsmath,amssymb,amsfonts}
\usepackage{algorithmic}
\usepackage{graphicx}
\usepackage{textcomp}
\usepackage{xcolor}
\usepackage{amsthm}
\usepackage{mathtools}
\newtheorem{observation}{Observation}
\usepackage{longtable}
\usepackage{booktabs}
\usepackage{array}
\usepackage{multirow}
\usepackage{tabularx,ragged2e,siunitx}
\usepackage{diagbox}
\usepackage{pifont}
\usepackage{float}
\usepackage{caption}
\usepackage{subfigure}
\usepackage{subcaption}
\usepackage{fancyhdr}
\usepackage[ruled,vlined,linesnumbered]{algorithm2e}
\usepackage{balance}
\usepackage{hyperref}
\usepackage{booktabs} 
\usepackage{optidef}
\usepackage{enumitem}
\usepackage{pifont}

\newtheorem{remark}{Remark}

\def\BibTeX{{\rm B\kern-.05em{\sc i\kern-.025em b}\kern-.08em
    T\kern-.1667em\lower.7ex\hbox{E}\kern-.125emX}}

\newcommand{\refsec}[1]{Section\@~\ref{sec:#1}}
\newcommand{\reffig}[1]{Fig\@.~\ref{fig:#1}}
\newcommand{\reftab}[1]{Table~\ref{tab:#1}}

\newcommand{\Comment}[1]{\textcolor{black}{#1}}

\begin{document}
\title{P3SL: Personalized Privacy-Preserving Split Learning on Heterogeneous Edge Devices}

\author{Wei~Fan,
        JinYi~Yoon,
        Xiaochang~Li,
        Huajie~Shao,
        and~Bo~Ji% <-this % stops a space
\IEEEcompsocitemizethanks{\IEEEcompsocthanksitem This research was supported in part by NSF grant CNS-2315851, the Commonwealth Cyber Initiative (CCI), and a Virginia Tech Presidential Postdoctoral Fellowship. A preliminary version of this work is to be presented as an invited paper at The 34th International Conference on Computer Communications and Networks (ICCCN 2025)~\cite{Icccn2025}.\protect\\
\IEEEcompsocthanksitem Wei Fan (fanwei@vt.edu), JinYi Yoon (jinyiyoon@vt.edu),
and Bo Ji (boji@vt.edu) are with the Department of Computer Science,
Virginia Tech, Blacksburg, VA. Xiaochang Li (xli59@wm.edu) and Huajie Shao (hshao@wm.edu) are with the Department of Computer Science,
William \& Mary, Williamsburg, VA.}% 
\thanks{Manuscript received April 19, 2005; revised August 26, 2015.}}

\markboth{Journal of \LaTeX\ Class Files,~Vol.~14, No.~8, August~2015}%
{Shell \MakeLowercase{\textit{et al.}}: Bare Demo of IEEEtran.cls for Computer Society Journals}

\IEEEtitleabstractindextext{%
\begin{abstract}
Split Learning (SL) is an emerging privacy-preserving machine learning technique that enables resource constrained edge devices to participate in model training by partitioning a model into client-side and server-side sub-models. While SL reduces computational overhead on edge devices, it encounters significant challenges in heterogeneous environments where devices vary in computing resources, communication capabilities, environmental conditions, and privacy requirements. Although recent studies have explored heterogeneous SL frameworks that optimize split points for devices with varying resource constraints, they often neglect personalized privacy requirements and local model customization under varying environmental conditions. To address these limitations, we propose P3SL, a Personalized Privacy-Preserving Split Learning framework designed for heterogeneous, resource-constrained edge device systems. The key contributions of this work are twofold. First, we design a personalized sequential split learning pipeline that allows each client to achieve customized privacy protection and maintain personalized local models tailored to their computational resources, environmental conditions, and privacy needs. Second, we adopt a bi-level optimization technique that empowers clients to determine their own optimal personalized split points without sharing private sensitive information (i.e., computational resources, environmental conditions, privacy requirements) with the server. This approach balances energy consumption and privacy leakage risks while maintaining high model accuracy. We implement and evaluate P3SL on a testbed consisting of 7 devices including 4 Jetson Nano P3450 devices, 2 Raspberry Pis, and 1 laptop, using diverse model architectures and datasets under varying environmental conditions. Experimental results demonstrate that P3SL significantly mitigates privacy leakage risks, reduces system energy consumption by up to 59.12\%, and consistently retains
high accuracy compared to the state-of-the-art heterogeneous SL system.
\end{abstract}

}

\maketitle
\IEEEdisplaynontitleabstractindextext
\IEEEpeerreviewmaketitle

\section{Introduction}
\IEEEPARstart{T}{he} rapid development of Internet-of-Things (IoT) devices has led to their integration into various aspects of daily life, performing tasks ranging from monitoring and sensing to machine learning (ML) and intelligent decision-making~\cite{ye2024asteroid,yao2017deepiot,thapa2022splitfed}. To protect data privacy, some research has proposed training entire machine learning models to process data locally~\cite{zhu2022device}. However, training entire ML models on resource-constrained edge devices presents significant challenges, including high energy consumption and prolonged training durations. To address these challenges, Split Learning (SL)~\cite{gupta2018distributed} has emerged as a promising solution. Unlike deploying the entire model on client side as in Federated Learning (FL)~\cite{heinbaugh2023data}, SL partitions an ML model into at least two sub-models: a client-side sub-model and a server-side sub-model. This division allows the client to process only a portion of the model, thereby reducing computational load and energy usage.

\begin{figure}[!t]
    \centering
    \includegraphics[width=0.99\linewidth]{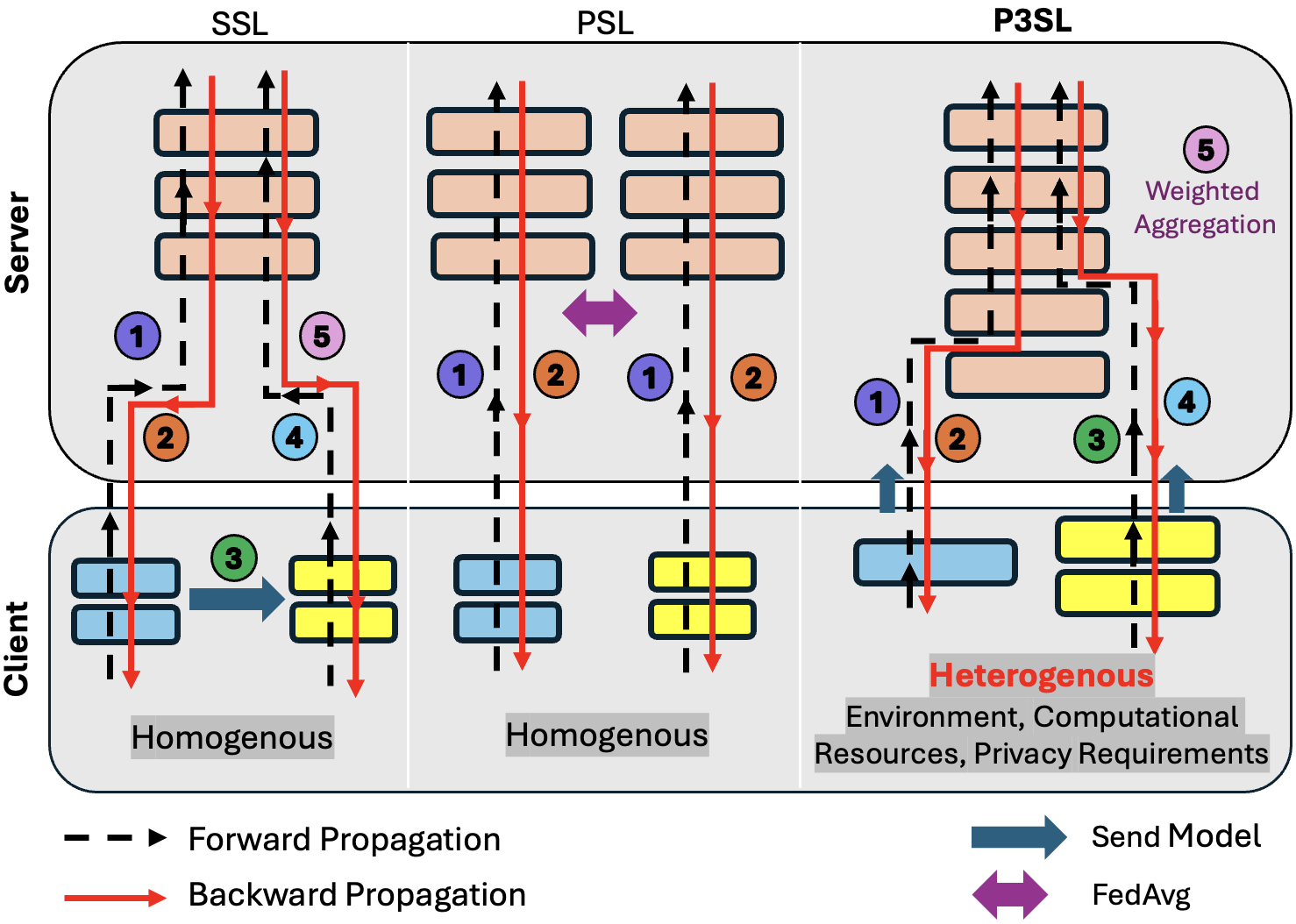}
    \caption{Comparison of SSL, PSL, and P3SL frameworks: SSL (left): Utilizes sequential training with homogeneous split points, requiring inter-client model sharing and lacking support for personalization. PSL (center): Enables parallel training but retains homogeneous split points, leading to high server resource costs and reduced accuracy due to the lack of personalization. \textbf{P3SL (right):} Allows personalized split points for edge devices, adapting to heterogeneous resource conditions including computational resources, privacy requirements, and environmental settings.}
    \vspace{-5mm}
    \label{fig:compare_SL}
\end{figure}

In real-world applications of SL, devices often operate with varying computational resources and privacy requirements in heterogeneous environments. However, most existing SL approaches assume a homogeneous client environment~\cite{sl1,sl2,sl3,sl4,sl5}, applying uniform split points across all devices. The split point in SL refers to the specific layer in a neural network where the model is divided between the client and the server, with the client processing layers before the split and the server handling layers after it. 

Sequential Split Learning (SSL)~\cite{gupta2018distributed} and Parallel Split Learning (PSL)~\cite{jeon2020privacy} are the two main homogeneous SL categories, as illustrated in Fig.~\ref{fig:compare_SL}. SSL achieves high accuracy with minimal server computation by training clients sequentially, but it requires inter-client model sharing, limiting model personalization and privacy. PSL addresses some limitations of SSL by allowing concurrent training without model sharing. However, PSL sacrifices accuracy due to missing client-side aggregation and requires significant server resources, making it unsuitable for large-scale edge deployments. These approaches fail to consider the diverse computational capabilities and privacy requirements of heterogeneous IoT ecosystems. For instance, a tablet in a common area and a Home Assistant device (e.g., Google Home) in a private space can collaborate to train a smart healthcare monitoring model under heterogeneous environmental conditions~\cite{Taylor2024-yp}. The tablet, with more computational resources, may process less sensitive data, while the Home Assistant device, with limited computational resources, handles highly sensitive private data. Applying a uniform SL strategy in such scenarios results in suboptimal performance, fails to address the unique requirements of each device, and provides inadequate privacy protection.

Existing works, such as ARES~\cite{SAMIKWA_2022}, ASL~\cite{li2024adaptivesplitlearningenergyconstrained}, and EdgeSplit~\cite{zhang2024resource}, have studied heterogeneous SL frameworks that optimize split points based on individual resource constraints. These frameworks focus on factors like training latency and computational resources for each client, offering an improvement over uniform SL strategies. However, these approaches often neglect several critical issues: (\romannumeral1) \emph{Privacy leakage across varying split points}: Different split points expose intermediate representations to varying levels of vulnerability~\cite{erdougan2022unsplit}, heightening the risk of data reconstruction attacks;
(\romannumeral2) \emph{Energy consumption and power constraints under heterogeneous environments}: Real-world deployments often occur in diverse environmental conditions, such as varying temperature and humidity, which can significantly impact total energy consumption and peak instantaneous power usage; 
(\romannumeral3) \emph{Personalized client models and personalized privacy requirements}: In practice, clients have varying privacy requirements based on their computing resources and environments.

%%--------table--------
\begin{table}[t]
\centering
\caption{Comparison of P3SL with existing SL methods}
\vspace{-1mm}
\label{tab:The-Comparison-of-Frameworks}
\setlength{\tabcolsep}{4pt}
\begin{tabular}{ccccc}%{p{1.75cm}p{1.0cm}p{1.1cm}p{1.8cm}p{1.3cm}}
\hline
{Methods} & {Energy} & {Privacy} & {Personalized Model}  & {Environments} \\ 
\hline
SSL~\cite{gupta2018distributed}   & \multicolumn{1}{c}{\ding{55}}   & \multicolumn{1}{c}{\ding{55}}  & \multicolumn{1}{c}{\ding{55}}  & \multicolumn{1}{c}{\ding{55}}  \\
ARES~\cite{SAMIKWA_2022}   & \multicolumn{1}{c}{\ding{51}}  & \multicolumn{1}{c}{\ding{55}}  & \multicolumn{1}{c}{\ding{55}}   & \multicolumn{1}{c}{\ding{55}} \\
ASL~\cite{li2024adaptivesplitlearningenergyconstrained}   & \multicolumn{1}{c}{\ding{51}}  & \multicolumn{1}{c}{\ding{55}}  & \multicolumn{1}{c}{\ding{55}}   & \multicolumn{1}{c}{\ding{55}} \\
EdgeSplit~\cite{zhang2024resource}   & \multicolumn{1}{c}{\ding{55}}  & \multicolumn{1}{c}{\ding{55}}  & \multicolumn{1}{c}{\ding{55}}   & \multicolumn{1}{c}{\ding{55}} \\
\textbf{P3SL (ours)}   & \multicolumn{1}{c}{\ding{51}}  & \multicolumn{1}{c}{\ding{51}} & \multicolumn{1}{c}{\ding{51}}  & \multicolumn{1}{c}{\ding{51}}  \\
\hline

\end{tabular}
\vspace{-3mm}
\end{table}

These limitations motivate us to propose \textbf{\underline{P}}ersonalized \textbf{\underline{P}}rivacy-\textbf{\underline{P}}reserving \textbf{\underline{S}}plit \textbf{\underline{L}}earning (P3SL), a novel SL framework designed for client devices to select optimal \textit{personalized} split points addressing varying privacy and energy needs in heterogeneous environmental conditions. A comparison of P3SL with the state-of-the-art frameworks is summarized in Table \ref{tab:The-Comparison-of-Frameworks}. Our P3SL addresses two main challenges below:

(\romannumeral1) How to effectively support numerous heterogeneous edge devices (clients) in SL to enable them to have private models with personalized split points while maintaining high accuracy. 

(\romannumeral2) How to determine the optimal split points and privacy protection (noise) levels for each edge device in heterogeneous environments without revealing sensitive information (e.g., their environments, computational resources, privacy requirements) to the server, while ensuring global model accuracy. 

To the best of our knowledge, \textsc{P3SL} is the first work towards personalized privacy-preserving SL for heterogeneous resource-constrained edge devices operating under varying environmental conditions. We summarize the main contributions as follows:
%\bo{anything to update to this paragraph?}
\begin{itemize}
    \item We propose \textsc{P3SL}, a novel framework incorporating a personalized sequential split learning pipeline that empowers clients to train personalized models while minimizing update costs and maintaining high model accuracy in heterogeneous, resource-constrained edge computing systems. This training pipeline directly addresses the first challenge by enabling clients to train sequentially with the server without inter-client model sharing. It provides enhanced privacy protection against data reconstruction attacks. Additionally, we deisgn a novel weighted aggregation technique to reduce the frequency of model updates sent to the server, significantly improving communication efficiency.
    
    \item To address the second challenge, we formulate the joint selection of split points and privacy protection levels for clients as a bi-level optimization problem. By employing a meta-heuristic approach~\cite{sinha_2020review, yao2024constrained}, we achieve an effective trade-off between energy consumption and privacy risks while maintaining high model accuracy. Notably, clients choose their split points and privacy protection levels according to their privacy preferences, without revealing sensitive information to the server.

    \item We implement \textsc{P3SL} on a real-world testbed consisting of seven edge devices, including four Jetson Nano P3450 devices, two Raspberry Pis, and a laptop. We evaluate its effectiveness using three model architectures (VGG16-BN, ResNet18, and ResNet101) across three datasets (Fashion-MNIST~\cite{Xiao2017FashionMNISTAN}, CIFAR-10~\cite{Krizhevsky2009LearningML}, and Flower-102~\cite{Nilsback08}) under various heterogeneous environmental conditions. Experimental results show that \textsc{P3SL} significantly enhances personalized privacy protection in terms of the Feature SIMilarity (FSIM)~\cite{Zhang_2011} score and against to Membership Inference Attacks (MIA)~\cite{MIA}, while reducing energy consumption by up to 59.12\% compared to existing SL systems and maintaining high global model accuracy. We further demonstrate robustness under large-scale settings and under dynamic network conditions, such as client disconnections and new client additions.
\end{itemize}

\section{Related Work}

\textbf{Split Learning Approaches.}
\Comment{Existing SL frameworks such as SSL~\cite{vepakomma2018split,gupta2018distributed} and PSL~\cite{jeon2020privacy,Pal2021ServerSideLG} assume uniform split points across all clients, which simplify implementation but fail to address the diverse resource or privacy constraints in heterogeneous environments. To address this, recent heterogeneity-aware SL approaches, such as ARES~\cite{SAMIKWA_2022}, ASL~\cite{li2024adaptivesplitlearningenergyconstrained}, HSFL~\cite{sun2024splitfederatedlearningheterogeneous}, and EdgeSplit~\cite{zhang2024resource}, dynamically determine split points based on each client's system-level factors, (e.g., computational latency, communication overhead, power consumption, and intermediate output size). However, these methods still lack mechanisms for personalized privacy protection, which is essential in settings where privacy risks vary across different clients. Furthermore, EdgeSplit and ASL are only validated through simulations, and while ARES is deployed on real-world edge devices, it relies on a high-capacity centralized server to handle all server-side computations, limiting its scalability under resource-constrained environments.}

\textbf{Defense Mechanisms against Data Reconstruction Attacks in SL.}
In split learning, attack approaches such as input data reconstruction~\cite{pasquini2021unleashingtiger,erdougan2022unsplit} can recover raw data from transmitted model parameters or intermediate representations, leading to privacy leakage problems. 
Several defense mechanisms have been proposed to mitigate these risks, including adding noise to raw data~\cite{khowaja2022get}, intermediate representation~\cite{li2022label,titcombe2021practical}, or model parameters~\cite{haim2022reconstructing,zhu2019deep}. Although these approaches can effectively mitigate data reconstruction attacks, the performance depends on factors such as the layer from which outputs are generated (e.g., outputs from deeper layers are harder to recover~\cite{erdougan2022unsplit}). Also, given the resource constraints, it is infeasible to fully commit to privacy protection without considering affordable computations in SL. \Comment{Existing methods that support heterogeneous split points among clients, such as ARES and SSL, do not offer privacy protection for clients with heterogeneous resources.}

\Comment{\textbf{Privacy‑Preserving Split Learning.} Several recent works aim to enhance privacy in split learning, but they do not support client-specific split point selection. For example, PPSFL~\cite{ZHENG2024231} incorporates private Group Normalization into each client’s model segment to protect against inter-client inference, but it requires all clients to use the same fixed split point. CURE~\cite{kanpak2024cureprivacypreservingsplitlearning} applies homomorphic encryption to server-side activations and estimates a global split point based on average client resources. However, it still enforces a uniform split point across all clients. P‑SL~\cite{NgocDuyPha2025} avoids weight sharing among clients and deploys separate server models for each client to reduce privacy leakage, but it does not allow coordination under a shared server architecture. Overall, the existing privacy-preserving SL methods lack the flexibility to support heterogeneous clients selecting their own split points and privacy requirement based on individual resource budgets and data sensitivity in a single training framework.}

To address these challenges, \textsc{P3SL} provides personalized privacy for each client, tailored to satisfy their computational requirements. In particular, by leveraging SSL, we ensure that privacy is preserved while maintaining high accuracy. Our approach enables each edge device to adapt to its heterogeneous resource, balancing between computational efficiency and robust privacy protection in practice.

\section{Motivation and key insights}%\shao{case studies}

\label{section:motivation}
In this section, we present case studies exploring the correlation between split points, energy consumption, and privacy leakage levels, motivating the design of
P3SL.

\subsection{Key Insights} 

To understand the impact of personalized split points on privacy leakage, energy consumption, and power constraints for each device, we implemented the data reconstruction attack of \emph{UnSplit}~\cite{erdougan2022unsplit} on the VGG16-BN~\cite{DBLP:journals/corr/SimonyanZ14a} model using the CIFAR-10~\cite{Krizhevsky2009LearningML} dataset. To measure the privacy leakage level, we use the FSIM~\cite{Zhang_2011} score, which measures the similarity between the original image and the reconstructed image. The FSIM score ranges from 0 to 1, where a higher FSIM score means a higher risk of privacy leakage.

\begin{figure}
    \centering
    \subfigure[Privacy leakage at three different training epochs $r$]{\includegraphics[width = 0.465\columnwidth]{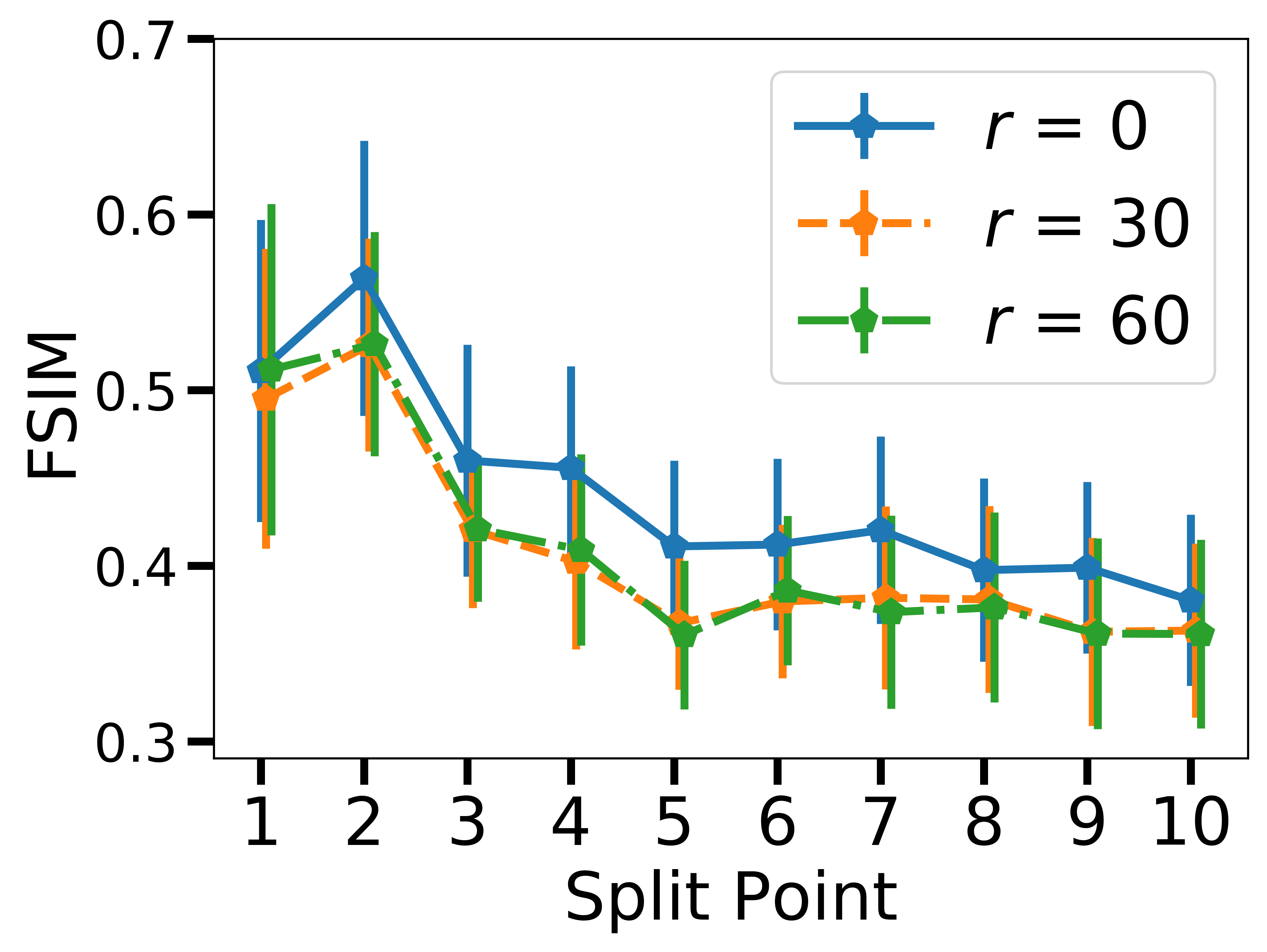}
    \label{fig:fsim-attack}}
    \subfigure[Defense with different variance $\sigma$ values for Laplacian noise, $Lap(0, \sigma^2)$ ]{\includegraphics[width = 0.465\columnwidth]{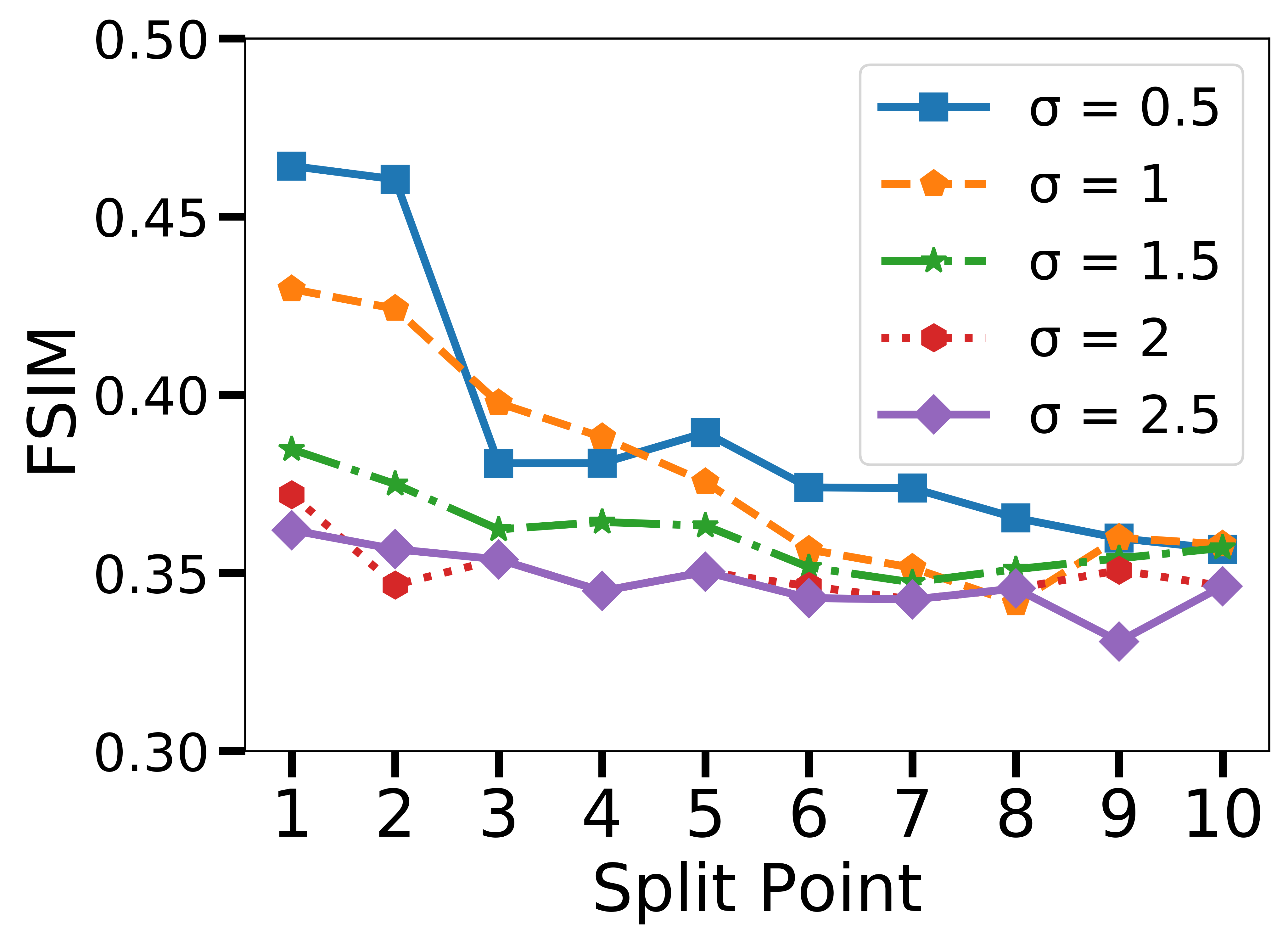}
    \label{fig:fsim-noise}}
    \vspace{-3mm}
    \caption{Impact of split points on privacy leakage and defense effectiveness}
    \vspace{-3mm}
    \label{fig:fsim-insight}
\end{figure}

\textbf{Privacy Leakage Across Split Points.}
First, we explore the privacy leakage across various split points by varying the number of training epochs of 0, 30, and 60 epochs (%corresponding to 0\%, 30\%, and 60\% of 
where the total 100 epochs required for training) as shown in \reffig{fsim-attack}. We examine the split points range from Layer 1 to Layer 10, with higher indices indicating that more layers are processed on the client side before sending intermediate outputs to the server. The results indicate that deeper split points result in lower FSIM scores, indicating better privacy protection as reconstructing input data from intermediate representations becomes more difficult. Furthermore, FSIM scores remain consistent across the three attack stages because the \textit{UnSplit} attack simultaneously recovers input data and inverts the client model. Consequently, the results of data reconstructed attack are not correlated to the different training stages.

\begin{observation}
As evidenced through FSIM scores, different split points present varying levels of privacy leakage; a shallower split point poses a higher risk of data reconstruction, necessitating aggressive noise addition.
\end{observation}

\textbf{Defense Effectiveness.} 
To mitigate data reconstruction attacks, we employed the noise addition approach inspired by \textit{NoPeekNN}~\cite{titcombe2021practical}, which injects Laplacian noise with zero mean and variance $\sigma^2$ to the intermediate representation. As shown in \reffig{fsim-noise}, we analyzed FSIM scores across different noise levels $\sigma$ ranging from 0.5 to 2.5 (incremented by 0.5). Our findings demonstrate that higher noise levels correlate with lower FSIM scores, indicating enhanced privacy protection. However, excessively high noise levels (e.g., $\sigma$ of 2.0 or 2.5) obscure distinctions among split points, resulting in similar FSIM scores across all split points. This highlights a trade-off: shallower layers have higher privacy leakage %as noted in 
(\textbf{Observation 1}) and thereby need greater noise injection, potentially hurting the model performance adversely.

\begin{observation}
To achieve a desirable privacy protection for each split point, it is essential to determine the minimum noise injection to reach the target privacy threshold.
\end{observation}

\begin{table}
    \centering
    \footnotesize
    \caption{The size of intermediate representations and the type of layer for split point 1 to 10 in VGG16-BN} 
    \label{tab:data-size}
\setlength{\tabcolsep}{5pt}
\begin{tabular}{ccc}
\hline

 \textbf{Split Point} & \textbf{Type of Layer} & \textbf{Data Size (MB)} \\ \hline
 1                & Conv 2D           & 33.56                 \\ 
 2                & BatchNorm 2D + ReLU           & 33.56     \\ 
 3                & Conv 2D           & 35.56                 \\
 4                & BatchNorm 2D + ReLU           & 35.56     \\
 5                 & Max pooling           & 8.39             \\ 
 6                & Conv 2D           & 16.78                 \\ 
 7                & BatchNorm 2D + ReLU           & 16.78     \\ 
 8                & Conv 2D           & 16.78                 \\ 
 9                & BatchNorm 2D + ReLU           & 16.78     \\ 
 10                  & Max Pooling          & 4.20             \\ 

\hline
\end{tabular}
\end{table}

%%%----
\begin{figure}
    \centering
    \subfigure[Energy consumption]{\includegraphics[width = 0.475\columnwidth]{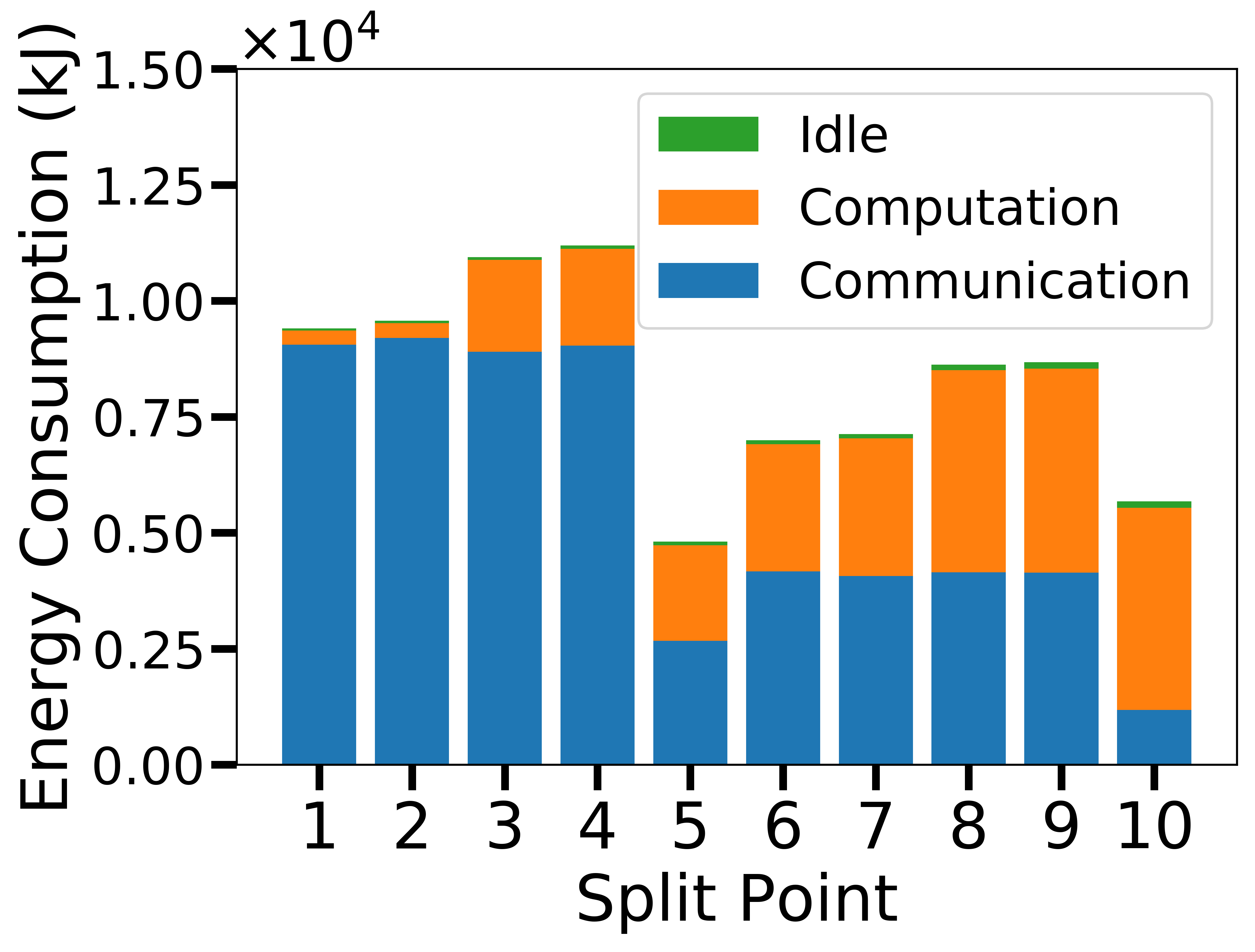}
    \label{fig:energy-consumption}}
    \subfigure[CDF of instantaneous power]{\includegraphics[width = 0.475\columnwidth]{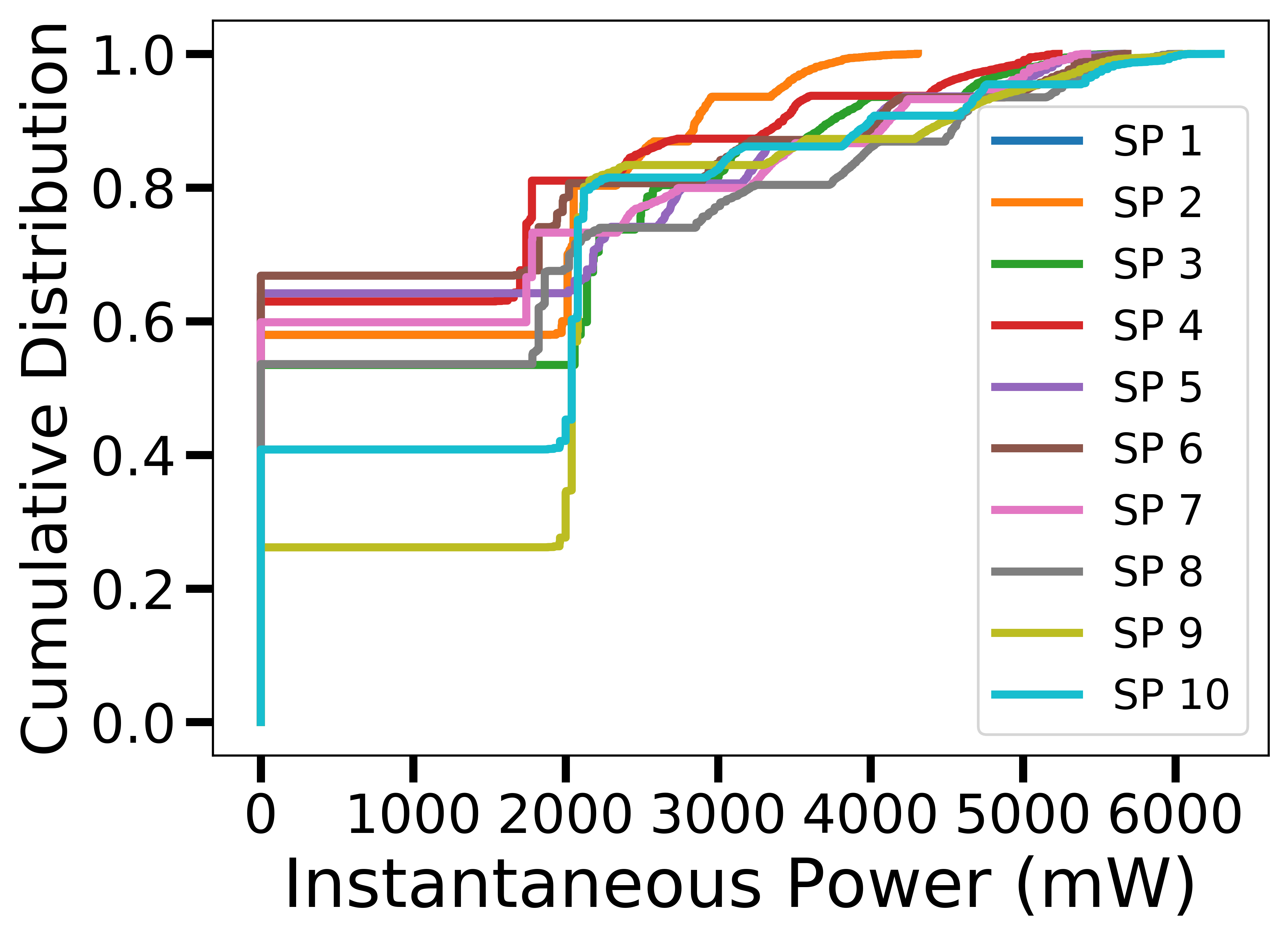}
    \label{fig:power-constraint}}
    \vspace{-3mm}
    \caption{Impact of different split points (SP) on energy consumption and instantaneous power}
    \label{fig:energy-power}
    \vspace{-3mm}
\end{figure}

\textbf{Energy Consumption and Power Constraints.} We evaluated the energy and power implications of split points in an SSL setting using four 4~GB NVIDIA Jetson Nano P3450 devices\cite{nvidia_jetson_nano}. \reftab{data-size} shows the size of intermediate representations at different split layers, and \reffig{energy-consumption} illustrates the average energy consumption across communication, computation, and idle states. Results indicate that communication energy consumption scales proportionally with the size of intermediate representations. Deeper split points, which generate smaller representations (i.e., lower data dimensions), reduce communication energy but increase computational energy due to more computation on the client side. In the same context, any layers with the same size of intermediate representation ideally have the same communication energy consumption, but deeper layers incur higher computational energy consumption. Therefore, total energy consumption increases with the deeper split points. In such cases, selecting the optimal split point with lower energy consumption is essential.

Furthermore, edge devices often face diverse environmental conditions, such as temperature variations, seasonal changes, or location-specific factors, which can cause overheating issues even in identical devices. \reffig{power-constraint} present the cumulative distribution of instantaneous power measured by the Jetson Nano devices, illustrating that deeper split points result in higher peak instantaneous power. To prevent overheating, it is crucial to ensure instantaneous power stays within the peak limit set for each client based on their environment conditions.

\begin{observation}
Deeper split points reduce communication energy but increase computational energy consumption, indicating that optimizing split points involves balancing these trade-offs to align with the power constraints of edge devices.
\end{observation}

These observations suggest a trade-off between privacy protection and %various 
energy/power consumption patterns across different split points: shallower split points increase (i) the risk of privacy leakage; (ii) communication energy; while reducing (iii) computational energy consumption.

\subsection{Design Goals}
Our design aims to achieve two important goals:
\begin{itemize}
    \item \textbf{Support Personalized Models.} Design a training pipeline that enables each client to maintain its personalized model, split point, and privacy protection (noise) level. This approach aims to accommodate resource-heterogeneous devices in collaborative settings.
    \item \textbf{Balance Energy Consumption and Privacy Leakage Risk.} Allow clients to customize split points and privacy protection levels to balance energy consumption and privacy leakage while ensuring high global model accuracy. Conduct an approach that effectively addresses these trade-offs to optimize overall performance across heterogeneous environments.
\end{itemize}
\section{P3SL Design} \label{sec:system}

We design P3SL with an SSL system (\reffig{P3SL System Architecture}) that allows edge devices to have heterogeneous split points along with personalized privacy protection in two ways: 1) personalized model to keep the private model locally; 2) personalized privacy protection to adjust the injected noise level. 
The clients can choose heterogeneous split points and personalized privacy protection levels on their intermediate representations before uploading them to the server. Besides, the global model is trained sequentially without requiring inter-client model sharing. Our system model is detailed in the following.

\subsection{P3SL Architecture}\label{sec:architecture}

Let $\mathcal{C}:=\{1, 2, \cdots, N\}$ denote a set of $N$ heterogeneous devices with varying computational capabilities. Each client $i \in \mathcal{C}$ has its local model $\boldsymbol{W}_{c_i}$ and a local dataset $\mathcal{D}_{i}$ consisting of $|\mathcal{D}_{i}|$ labeled samples. The local dataset is defined as $\mathcal{D}_{i} := \{({x}_{i,j}, {y_{i,j}})\}_{j=1}^{|{\mathcal{D}_i}|}$, where ${x}_{i,j}$ is the $j$-th input data sample, and $y_{i,j}$ is its corresponding label. Hence, the global dataset $\mathcal{D}$ is denoted as $\mathcal{D}
= {\bigcup_{i\in\mathcal{C}}}\mathcal{D}_i$. 

For simplicity, we define $\boldsymbol{W}^{a:b} \coloneqq {[\boldsymbol{W}^{a}, \boldsymbol{W}^{a+1}, \cdots, \boldsymbol{W}^{b}]}$ as the model containing layers from $a$ to $b$. In \textsc{P3SL}, all clients collaborate to train the global model $\boldsymbol{W} := \boldsymbol{W}^{1:k}$, where $k$ is the total number of layers in the global model. The global model $\boldsymbol{W}$ serves as the shared server model and is partitioned at a split point $s_i$ into two sub-models. Specifically, for client $i$, the first $s_i$ layers are located on the client side, so the client model is denoted as $\boldsymbol{W}_{c_i} \coloneqq \boldsymbol{W}_{c_i}^{1:s_i}$. The remaining layers, from $s_i+1$ to $k$, are located on server side for client $i$, and are denoted as $\boldsymbol{W}_{g_i} := \boldsymbol{W}_{g_i}^{s_i+1:k}$. In addition, the server determines the maximum allowable split point $s_{\text{max}}$ for all clients. Each client $i$ can choose their split point $s_i \in \{1, 2, \dots, s_{\text{max}}\}$ based on local privacy requirements and energy consumption constraints. By definition, $1 \leq s_i \leq s_{\text{max}} \leq k$.  

\begin{figure}[!t]
    \centering
    \includegraphics[width=0.99\linewidth]{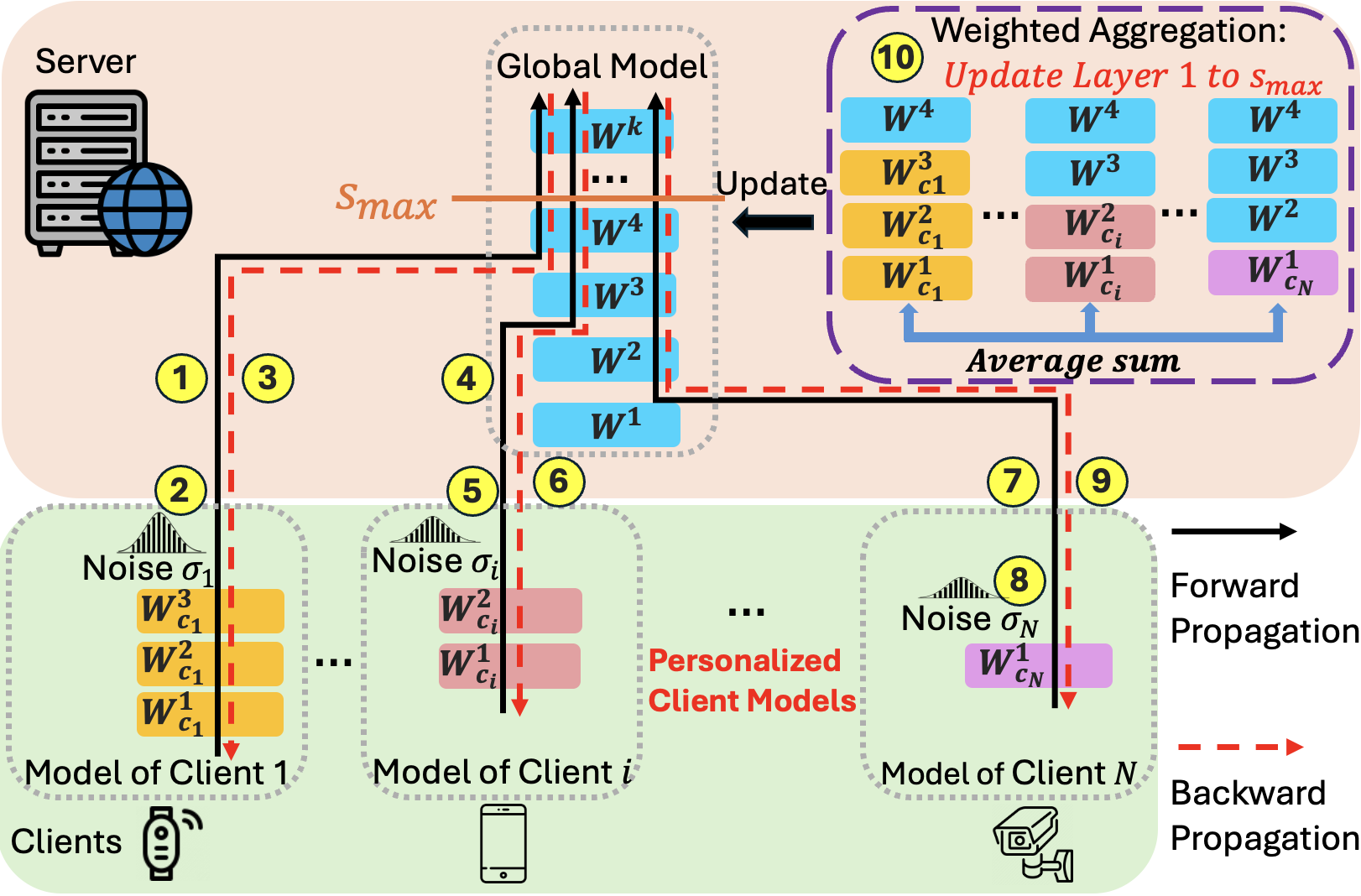}
    \caption{System architecture of \textsc{P3SL}: with 
    $\boldsymbol{s}$ of split points and $\boldsymbol{\sigma}$ of privacy protection (noise) levels, clients generate intermediate representations, inject noise, and upload them to the server for {\large \textcircled{\small 1}}-{\large \textcircled{\small 9}} sequential model training. Every $R$ epoch, clients upload their local models to the server for {\large \textcircled{\small 10}} weighted aggregation. The process {\large \textcircled{\small 1}}-{\large \textcircled{\small 10}} repeats until the global model converges.} 
    \label{fig:P3SL System Architecture}
\end{figure}

The overall training procedure of personalized sequential split learning in P3SL involves the following steps, illustrated in \reffig{P3SL System Architecture}:

(\romannumeral1) \textit{Forward Propagation on Client-Side Model:} After initializing the global model at the server side, P3SL sequentially selects each client for model training. The selected $i$-th client samples local labeled data $(x_{i,j},y_{i,j}) \in \mathcal{D}_{i}$ and forward propagates the input data ${x}_{i,j}$ through its local model to generate intermediate representations $\hat{z}_{i,j}:= g({x}_{i,j}\mid\boldsymbol{W}_{c_i})$. Here, $g(x \mid \boldsymbol{W})$ is the mapping function between input data $x$ and its predicted value $\hat{z}$ given model $\boldsymbol{W}$ (see {\large \textcircled{\small 1}{\large \textcircled{\small 4}}{\large \textcircled{\small 7}}}).

(\romannumeral2) \textit{Privacy-Preserving Intermediate Representation Transmission:} The client $i$ injects Laplacian noise $\eta_{i,j}$ drawn from $Lap(0, {\sigma_i}^2)$ into the intermediate representation $\hat{z}_{i,j}$, where $\sigma_i$ is the personalized privacy protection (noise) level. Then, both noise-injected representation $\hat{z}_{i,j}+ \eta_{i,j}$ and the corresponding label $y_{i,j}$ will be sent to the server (see {\large \textcircled{\small 2}{\large \textcircled{\small 5}}{\large \textcircled{\small 8}}}). 

(\romannumeral3) \textit{Forward Propagation on Server-Side Model:} The server receives the privacy-protected intermediate representation, and then processes it through the remaining layers of the global model $\boldsymbol{W}_{g_i}:=\boldsymbol{W}^{{s_{i}}+1:k}$, and generate the final prediction $\hat{y}_{i,j}$.

(\romannumeral4) \textit{Gradient Calculation and Backward propagation:} The server computes the gradients based on the loss function for each client $i$, which is denoted as $\mathcal{L}(\hat{y}_{i,j}, {y}_{i,j} \mid \boldsymbol{W}_{c_i} \oplus \boldsymbol{W}_{g_i})$, where $\oplus$ represents model concatenation. For example, the concatenation of models $\boldsymbol{W}^{a:b}$ and $\boldsymbol{W}^{c:d}$ is defined as $\boldsymbol{W}^{a:b} \oplus \boldsymbol{W}^{c:d} := [\boldsymbol{W}^{a}, \boldsymbol{W}^{a+1},\cdots, \boldsymbol{W}^{b}, \boldsymbol{W}^{c}, \boldsymbol{W}^{c+1}, \cdots \boldsymbol{W}^{d}]$. During backpropagation, the server only updates the layer in $\boldsymbol{W}_{g_i}$ and sends the gradients back to the client. Then, the client completes the backward propagation and updates its model parameters (see {\large \textcircled{\small 3}{\large \textcircled{\small 6}}{\large \textcircled{\small 9}}}). 

(\romannumeral5) \textit{Model Aggregation and the Global Model Updates:} After training for $R$ epochs, clients upload their local model parameters to update the global model's first $s_{\text{max}}$ layers $\boldsymbol{W}^{1 : s_{\text{max}}}$. All uploaded parameters $\boldsymbol{W}_{c_i}$ will be aggregated using Eq.~\eqref{eq:fed_avg}. For clients missing layers from $s_{i}+1$ to $s_{\text{max}}$, the corresponding global model layers are used to fill in during aggregation. The global model $\boldsymbol{W}$ is then updated as follows: 
\begin{equation}
    \label{eq:fed_avg}
    \boldsymbol{W} = (\frac{1}{N}\sum_{i=1}^{N} (\boldsymbol{W}_{c_i} \oplus \boldsymbol{W}^{s_{i}+1 : s_{\text{max}}})) \oplus \boldsymbol{W}^{s_{\text{max}}+1 : k}.
\end{equation}
This weighted model aggregation technique ensures that $\boldsymbol{W}^{1 : s_{\text{max}}}$ is not overly influenced by a few clients with larger $s_i$ during the aggregation process. Note that the aggregated model is not distributed in order to preserve model personalization for each client (see {\Large \textcircled{\small 10}}). 

This training process repeats until all clients complete their updates and the system loss function is minimized.

\subsection{Profiling} \label{sec:profiling}
In this subsection, we present the profiling process for the P3SL framework, which includes profiling the \textit{Privacy Leakage Table} on the server side and the \textit{Energy and Power Consumption Table} on the client side. These tables are essential to formulating and solving the bi-level optimization problem (Section \ref{sec:problem-solution}). In addition, the server decides the minimum accuracy threshold $A_{\text{min}}$ in order to ensure the performance of global model.

\begin{figure}[!t]
    \centering
    \subfigure[\textit{Privacy Leakage Table} profiled by server]{\includegraphics[width = 0.45\columnwidth]{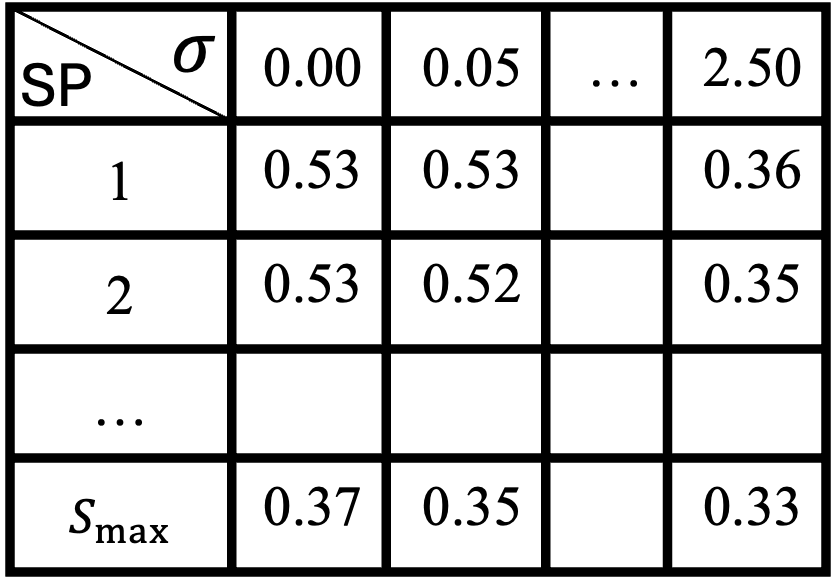}
    \label{fig:privacy-table}}
    \subfigure[\textit{Energy and Power Consumption Table} profiled by each client $i$]{\includegraphics[width = 0.47\columnwidth]{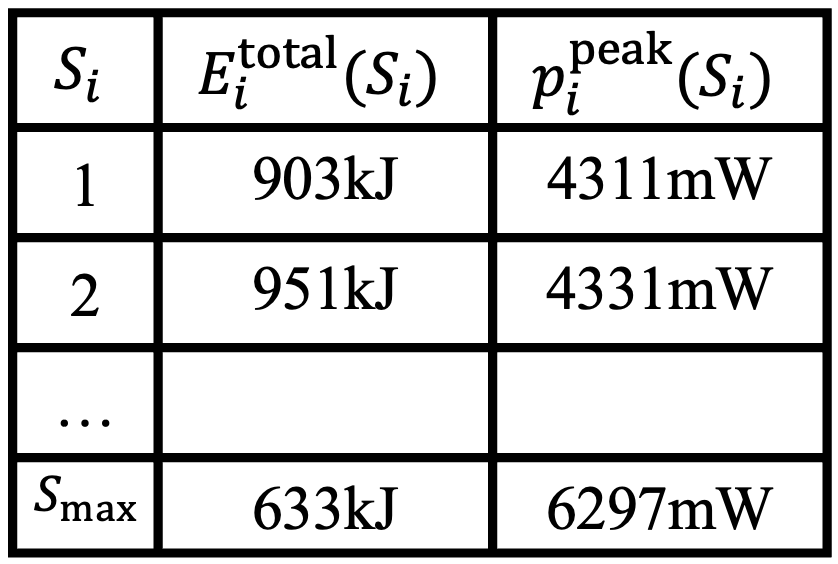}
    \label{fig:energy-tables}}
    \vspace{-1mm}
    \caption{An example of profiling tables}
    \vspace{-4mm}
    \label{fig:profiled-tables}
\end{figure}

\textbf{Constructing \textit{Privacy Leakage Table}.} We assume the model architecture and the intermediate representations from clients are accessible to an attacker. As shown in \reffig{fsim-attack}, the effectiveness of the data reconstruction attack is consistent across different training stages, and thus we employ the attack before training~\cite{erdougan2022unsplit}. Privacy leakage risk is assessed using the FSIM score~\cite{Zhang_2011}, which measures the similarity between the original input data and reconstructed data. Since local clients collaboratively train the same global model, the server simulates data reconstruction attacks on a public dataset to generate the \textit{Privacy Leakage Table}. As shown in \reffig{privacy-table}, this table evaluates FSIM for each split point from 1 to $s_{\text{max}}$ within a specified noise range (0.00 to 2.50 with 0.05 interval) and is then distribute to each client.

\textbf{Constructing \textit{Energy and Power Consumption Table}.} 
Each client profiles energy consumption and peak instantaneous power for different split points. For each split point $s_i$ from 1 to $s_{\text{max}}$, the $i$-th client generates an \textit{Energy and Power Consumption Table}, as shown in \reffig{energy-tables}. This table includes: (\romannumeral1) the average total energy consumption, $E_i^{\text{total}}(s_i)$; and (\romannumeral2) peak instantaneous power, $p_i^{\text{peak}}(s_i)$. Meanwhile, the client tests out its (\romannumeral3) maximum device power threshold $P_i^\text{max}$ to prevent overheating. 

\textbf{Minimum Accuracy Threshold Determination.} \label{sec:minimum-accuracy}
When initiating the training setup for \textsc{P3SL}, the server determines the minimum accuracy threshold, $A_\text{min}$, which is the required global model accuracy for completing the optimization process. To establish this threshold, the server simulates the training process without noise injection using a publicly available dataset that closely matches the distribution of the clients' datasets. This simulation provides the reference accuracy $A_{\text{ref}}$, as an ideal accuracy baseline. During this phase, the server also fine-tunes the training hyperparameters and shares them with clients to prepare for the following SL training. The minimum accuracy threshold $A_\text{min}$ is then calculated as:
\begin{equation}
\label{eq:acc-threshold}
    A_{\text{min}} = \beta \cdot A_{\text{ref}},
\end{equation}
where $\beta$ is a discount factor set by the server to define the acceptable accuracy sacrifice for enhanced privacy protection. To ensure consistency, $\beta$ is set by the server at the beginning of the process and remains fixed throughout the training. 

\section{Problem Formulation and Proposed Solution}\label{sec:problem-solution}
In real-world scenarios, clients are often reluctant to share sensitive information with the server due to privacy concerns. If the server independently determines split points and privacy protection levels for all clients, it necessitates the disclosure of sensitive client-specific information (e.g., environmental conditions, computational resources, and privacy requirements). To address this challenge, \textsc{P3SL} facilitates joint decision-making between the server and clients, optimizing overall system performance.

In this section, we formulate a bi-level optimization problem (\refsec{bilevel-formulation}) and propose a solution (\refsec{bi-level}) that allows the server and clients to jointly determine the optimal privacy protection levels and split points. This solution effectively balances the trade-off between privacy protection and energy efficiency while maintaining high global model accuracy. 

\subsection{Bi-Level Optimization Problem Formulation}\label{sec:bilevel-formulation}
Let us first define two decision variables: the privacy protection level vector $\boldsymbol{\sigma}$ and the split point vector $\boldsymbol{s}$. Denote $\boldsymbol{\sigma}:=[\sigma_{1}, \sigma_{2}, \cdots, \sigma_{N}]$, where $\sigma_i$ represents the privacy protection (noise) level for $i$-th client. Similarly, denote $\boldsymbol{s}:=[s_1, s_2, \cdots, s_N]$, where $s_i$ represents the split point chosen by $i$-th client.

For the lower-level optimization, after receiving the upper-level decision on the privacy protection level $\sigma_i$ from the server, client $i$ determines its own optimal split point $s_i$ by minimizing the local objective function $f$. Since minimizing total energy consumption $E_i^{\text{total}}(s_i)$ and privacy leakage $\text{FSIM}(\sigma_i, s_i)$ are potentially conflicting, $f$ is formulated as their weighted sum. The total energy consumption $E_i^{\text{total}}(s_i)$ is obtained from the \textit{Energy and Power Consumption Table} (\reffig{energy-tables}), while the privacy leakage $\text{FSIM}(\sigma_i, s_i)$ is obtained from \textit{Privacy Leakage Table} (\reffig{privacy-table}). The relative importance of $E_i^{\text{total}}(s_i)$ and $\text{FSIM}(\sigma_i, s_i)$ for the $i$-th client is determined by the personalized privacy sensitivity coefficient $\alpha_i \in \left[ 0,1 \right] $, which reflects client's preference between total energy consumption and privacy leakage. A higher $\alpha_i$ indicates a stronger preference for privacy protection, leading to increased noise injection and deeper split points. Conversely, a lower $\alpha_i$ emphasizes energy efficiency, favoring reduced energy consumption with shallower split points. The local objective function $f$ for each client $i$ is formulated as:
\begin{equation}\label{eq:total_cost_function}
    f(\sigma_i, s_i) := \alpha_i \cdot \text{FSIM}(\sigma_i, s_i) + (1-\alpha_i) \cdot E_i^{\text{total}}(s_i). 
\end{equation}

To optimize the privacy across all clients while maintaining high global model accuracy, the server (upper-level) decides the privacy protection level vector $\boldsymbol{\sigma}$ to minimize the total privacy leakage $\text{FSIM}(\sigma_i, s_i)$ across all clients. Meanwhile, the server ensures that the global model accuracy $G_{\text{acc}}(\boldsymbol{\sigma},\boldsymbol{s})$ remains above the minimum accuracy threshold $A_{\text{min}}$. This bi-level optimization problem is formulated as:
\begin{subequations} \label{eq:bilevel}
\begin{align}
    \label{eq:upper_objective} \min_{\boldsymbol{\sigma}, \boldsymbol{s}}& ~F(\boldsymbol{\sigma}, \boldsymbol{s}) := \sum_{i=1}^{N} \text{FSIM}(\sigma_i, s_i) & \\ \label{eq:constraint 1}
    \text{s.t.} &  ~s_i \in \arg\min_{s_i} f(\sigma_i, s_i), \forall{i} \in \mathcal{C},\\ \label{eq:constraint 2}
     &~G_{\text{acc}}(\boldsymbol{\sigma}, \boldsymbol{s}) \geq A_{\text{min}}, \\ \label{eq:constraint 3}
     &  ~p_i^{\text{peak}}(s_i) \leq P_i^\text{max}, \forall{i} \in \mathcal{C}.  
\end{align}
\end{subequations}
Here, Eq.~\eqref{eq:constraint 1} specifies that each client $i$ determines its optimal split point $s_i$ by minimizing local objective function $f$. Eq.~\eqref{eq:constraint 2} guarantees that the global model accuracy $G_{\text{acc}}(\boldsymbol{\sigma},\boldsymbol{s})$ remains above the minimum accuracy threshold $A_{\text{min}}$. Finally, Eq.~\eqref{eq:constraint 3} ensures that the peak instantaneous power $p_i^{\text{peak}}(s_i)$ for client $i$ does not exceed the maximum device power threshold $P_i^\text{max}$ to prevent overheating. The threshold $P_i^\text{max}$ is obtained from the client's energy and power consumption profiling process (\refsec{profiling}). 

\subsection{Proposed Solution}\label{sec:bi-level}
We employ a meta-heuristic algorithm~\cite{sinha_2020review} to solve the bi-level optimization problem. The upper-level optimization problem, defined in Eq.~\eqref{eq:upper_objective}, focuses on optimizing privacy across all clients while ensuring the global model accuracy meets the minimum threshold. The lower-level optimization problem, described in Eq.~\eqref{eq:total_cost_function}, aims to optimize the trade-off between total energy consumption and privacy leakage for each client. These upper-level and lower-level optimization problems are solved sequentially, enabling iterative refinement and convergence toward an overall optimal solution~\cite{Talbi2013}. The optimization process begins with the server (upper-level) generating an initial solution based on the upper-level objective function and distributing it to the clients (lower-level). Each client then determines its split point by minimizing the local objective function and sends its solution back to the server. This nested process continues iteratively until the overall optimization converges.

We implement this approach in the P3SL framework through three key steps:

\textit{(\romannumeral1) Initial Upper-Level Solution.} The server initiates the process by generating a \textit{Noise Assignment Table} based on the \textit{Privacy Leakage Table}. This \textit{Noise Assignment Table} specifies the privacy protection (noise) levels corresponding to each split point, ensuring low privacy leakage with robust privacy protection. The table is then distributed to the clients, allowing them to select their optimal split points.

\textit{(\romannumeral2) Initial Lower-Level Solution.} First, each client $i$ determines its personalized privacy sensitivity coefficient $\alpha_i \in \left[ 0,1 \right] $. Next, the client $i$ identifies the deepest split points $s_{\text{max}}^{(c_i)}$, ensuring no overheating, and the split point $s_{\text{min}}^{(c_i)}$ that offers the lowest total energy consumption. If energy consumption increases with deeper layers, client $i$ selects the optimal split point from 1 to $s_{\text{max}}^{(c_i)}$. Conversely, if energy consumption decreases, the client selects the optimal split point between $s_{\text{min}}^{(c_i)}$ to $s_{\text{max}}^{(c_i)}$. Finally, clients select their optimal split points by minimizing the local objective function (Eq.~\eqref{eq:total_cost_function}), using the \textit{Noise Assignment Table} provided by the server. Afterward, the client sends its optimal split point selection $s_i$ to the server.

\textit{(\romannumeral3) Iterative Optimizing.} The server constructs $\boldsymbol{s}$ and $\boldsymbol{\sigma}$ to perform sequential training with clients and compute the global model accuracy $G_{\text{acc}}(\boldsymbol{\sigma}, \boldsymbol{s})$. If the global model accuracy is greater or equal to $A_{\text{min}}$, the optimization process is finished. Otherwise, the server updates the \textit{Noise Assignment Table} using noise reassignment strategies and redistributes it to clients. The clients then re-optimize their split points as described in (\romannumeral2), and the process repeats until the global accuracy reaches $A_{\text{min}}$.

\textbf{Initial Noise Assignment Table.}
The generation of the initial \textit{Noise Assignment Table} starts by defining an FSIM threshold, $T_{\text{FSIM}}$, to ensure that reconstructed data cannot be accurately classified. Specifically, the server uses a well-trained model to evaluate the reconstructed data with different levels of noise injection. Along with the \textit{Privacy Leakage Table}, the server identifies $T_{\text{FSIM}}$ at which the classification accuracy falls below $\frac{1}{N_\text{class}}$, where $N_\text{class}$ denotes the number of classes. 

Based on this analysis result, the server creates the \textit{Noise Assignment Table} by selecting the minimum privacy protection (noise) level for each split point that ensures the FSIM score remains below the threshold $T_{\text{FSIM}}$. This table is then distributed to all clients to initialize their optimal split point selection.

\textbf{Noise Reassignment Strategies.}
If the global accuracy, $G_{\text{acc}}(\boldsymbol{\sigma}, \boldsymbol{s})$, falls below the minimum accuracy threshold $A_{\text{min}}$, noise levels must be reduced to improve the global model's utility. To achieve this, the server adjusts the noise assignments and distributes an updated \textit{Noise Assignment Table} to each client. Denote global model accuracy at round $t$ as $A^{t}$, and the privacy protection (noise) level assigned for client $i$ at round $t$ as $\sigma_i^t$. For the next round $t+1$, the server calculates the new noise level $\sigma_i^{t+1}$ for each split point as:

\begin{equation}
    \sigma_i^{t+1} = \sigma_i^t \times (1-2 \cdot (A_{\text{min}} - A^{t})).
\end{equation}
This equation is designed such that a larger difference between $A^{t}$ and $A_{\text{min}}$ results in a greater reduction in privacy protection (noise) level for each client, thereby improving the global model's accuracy.

\section{Evaluation}

In this section, we aim to address several key research questions (RQs) on the privacy protection, energy efficiency, adaptability, and overall performance of P3SL, through comprehensive evaluations.
\begin{itemize}
    \item RQ1: How does \textsc{P3SL} compare to baseline methods in terms of privacy leakage, energy consumption, and global model accuracy under heterogeneous environmental conditions?
    \item RQ2: How flexible is \textsc{P3SL} in adapting to varying environmental conditions, such as temperature and cooling settings, and maintaining consistent performance?
    \item RQ3: How effective is \textsc{P3SL} for personalized privacy protection for individual clients with diverse resource constraints and privacy requirements?
    \item \Comment{RQ4: How adaptable is \textsc{P3SL} under dynamic network conditions, such as client disconnections or new client additions, while consistently preserving system accuracy and stability?}
    \item \Comment{RQ5: How does a large number of devices impact the accuracy and privacy protection performance of \textsc{P3SL}?}
    \item \Comment{RQ6: How robust is \textsc{P3SL} against membership inference attacks, and how effectively does it mitigate membership information leakage?}
\end{itemize}

\subsection{Implementation and Deployment}
\textbf{Device Settings.} We implement \textsc{P3SL} on four 4~GB NVIDIA Jetson Nano P3450 devices, two 4~GB Raspberry Pi, and one Laptop without GPU for the clients. The central server is equipped with an AMD Ryzen Threadripper PRO 5995WX @ 2.7~GHz, 128~GB RAM, and two NVIDIA GeForce RTX 4090 GPUs.
% \textbf{Implementation Details.}
We use \texttt{WebSocket} API to establish network connections and transmission components for the Jetson Nano devices. We utilize Kasa system monitoring tool~\cite{Kasa} to measure the real-time power consumption of all devices.

\begin{table}[!t]
    \centering
    \footnotesize
    \setlength{\tabcolsep}{3pt}
    \caption{Two environment condition settings and corresponding allowable deepest split point}
    \vspace{-3mm}
    \label{tab:diff-env-sp}    
    \subtable[Environment condition settings]{
        \centering
        \begin{tabular}{cccccc}
            \hline
            \multirow{2}{*}{\textbf{Client ID}} & \multicolumn{2}{c}{\textbf{Environment Setting A}} && \multicolumn{2}{c}{\textbf{Environment Setting B}} \\ \cline{2-3} \cline{5-6} %\cmidrule(lr){2-3}\cmidrule(lr){4-5}
            
            & \textbf{AC Temp.} & \textbf{Cooling Fan} & & \textbf{AC Temp.} & \textbf{Cooling Fan} \\
            \hline
            1 (Jetson Nano) & 30 $^{\circ} C$ & OFF && 30 $^{\circ} C$ & ON \\
            2 (Jetson Nano) & 30 $^{\circ} C$ & ON && 20 $^{\circ} C$ & OFF \\
            3 (Jetson Nano) & 20 $^{\circ} C$ & OFF && 15 $^{\circ} C$ & OFF \\
            4 (Jetson Nano) & 20 $^{\circ} C$ & ON && 15 $^{\circ} C$ & ON \\
            5 (Raspberry Pi) & 20 $^{\circ} C$ & OFF && 20 $^{\circ} C$ & OFF \\
            6 (Raspberry Pi) & 20 $^{\circ} C$ & ON && 20 $^{\circ} C$ & ON \\
            7 (Laptop) & 20 $^{\circ} C$ & AUTO & & 20 $^{\circ} C$ & AUTO \\
            \hline
        \end{tabular}%
        \label{tab:env-condition}
    }

\vspace{-1mm}
    \subtable[Allowable deepest split point $s_{\text{max}}^{(c_i)}$ avoid overheating for each client]{
    \centering
        \begin{tabular}{cccccc}
            \hline
            \multirow{2}{*}{\textbf{Client ID}} & \multicolumn{2}{c}{\textbf{Environment Setting A}} && \multicolumn{2}{c}{\textbf{Environment Setting B}} \\ \cline{2-3} \cline{5-6} 
            
            & \textbf{VGG16-BN} & \textbf{ResNet} && \textbf{VGG16-BN} & \textbf{ResNet} \\
            \hline
            1 (Jetson Nano) & 2 & 2 && 4 & 4 \\
            2 (Jetson Nano) & 4 & 3 && 7 & 7 \\
            3 (Jetson Nano) & 7 & 6 && 8 & 10 \\
            4 (Jetson Nano) & 10 & 10 && 10 & 10 \\
            5 (Raspberry Pi) & 1 & 1 && 1 & 1 \\
            6 (Raspberry Pi) & 2 & 2 && 2 & 2 \\
            7 (Laptop) & 10 & 10 && 10 & 10 \\
            \hline
        \end{tabular}%
        \label{tab:deep-sp}
    }
    \vspace{-3mm}
\end{table}

\textbf{Environment Settings.}
We conduct experiments in two controlled environmental settings. For each setting, we place each device in different rooms with varying environmental room temperatures and toggled the cooling fans of the devices. For the laptop (Client 7), the cooling system operates automatically, so its fans were not manually controlled. The detailed environmental conditions are described in \reftab{env-condition}. In order to address potential overheating issues, we measure the deepest (i.e., maximum) split points, $s_{\text{max}}^{(c_i)}$,  by monitoring the instantaneous power consumption of each client, as shown in \reftab{deep-sp}.

\Comment{\textbf{Sleep-Awake Scheduling.} P3SL adopts sleep-awake scheduling~\cite{Rehman_sleep_awake} to reduce idle energy consumption by recording energy usage only when the devices are in \textit{awake} mode. For exmaple, for Jetson Nano devices' idle states, instead of maintaining \textit{awake} to consume around 1.8~watts, the devices are set to \textit{sleep} mode, consuming nearly 0~watts. Therefore, energy consumption during \textit{sleep} periods is excluded, and only energy consumption during the \textit{awake} periods will be considered in the total energy consumption.}

\textbf{Datasets.}
We evaluate the system using three benchmark datasets: 
(\romannumeral1) CIFAR-10~\cite{Krizhevsky2009LearningML}, (\romannumeral2) Fashion-MNIST~\cite{Xiao2017FashionMNISTAN}, and (\romannumeral3) Flower-102~\cite{Nilsback08}.

% \begin{figure}
%     \centering
%     \includegraphics[width=\linewidth]{figure/architecture figures.png}
%     \caption{Model architecture of (a) ResNet18 and (b) VGG16-BN \shao{we do not need this figs}} 
%     \label{fig:model-arch}
% \end{figure}

\textbf{Model Architecture and Personalized Settings.}
We leverage three different models from lightweight to large-scale: ResNet18 (11M), ResNet101 (43M), and VGG16-BN (135M), with the deepest split point, $s_{\text{max}}$, set to 10. Each client uses a batch size of 256 and a learning rate of 0.01. Models are aggregated every 5 epochs. The personalized privacy sensitivity coefficient $\alpha_i$ varies across clients to reflect differing preferences between energy consumption and privacy leakage costs. Specifically, we set $\alpha_1 = 0.4, \alpha_2 = 0.2, \alpha_3 = 0.5, \alpha_4 = 0.9, \alpha_5 = 0.7, \alpha_6 = 0.3, \alpha_7 = 0.8 $. Additionally, the discount factor is set to $\beta=5\%$, representing the maximum tolerated accuracy sacrifice for the global model, to enhance clients' privacy protection.

\textbf{Evaluation Metrics.}
We evaluate the system performance based on three main metrics: (\romannumeral1) overall and personalized privacy leakage across clients; (\romannumeral2) energy consumption; and (\romannumeral3) accuracy. The overall system privacy leakage, denoted as $\text{FSIM}_{\text{total}}$, is to reflect the whole system's privacy leakage level, measured by averaging the sum of all clients' privacy leakage index values over five rounds. We further investigate each client's personalized privacy leakage $\text{FSIM}$. 
A lower FSIM score indicates better privacy protection, with even a small improvement (e.g., 0.025) in FSIM representing a significant reduction in privacy leakage~\cite{Zhang_2011}. 
Energy consumption is measured as $\overline{E}_{{\text{total}}}$, representing the average energy consumption per epoch by all edge devices over five test rounds. This includes energy consumption for communication, computation, and idle modes. Lastly, accuracy refers to the global model's testing accuracy for assessing the model utility.
 
\textbf{Baselines.}
We compare \textsc{P3SL} with two state-of-the-art approaches as baselines: (\romannumeral1) \textit{ARES}~\cite{SAMIKWA_2022}, which employs heterogeneous split points in PSL; (\romannumeral2) \textit{Sequential Split Learning (SSL)}~\cite{gupta2018distributed}, a classic SL framework that enforces homogeneous split points among all clients. In SSL, model parameters must be transmitted to initialize adjacent clients' training, making it difficult to support heterogeneous split points. Since neither \textit{ARES} nor \textit{SSL} considers for privacy leakage, we ensure a fair comparison by applying the same noise injection with the same variance $Lap(0, \sigma^2=2.5^2)$ as used in \textsc{P3SL} to all baselines.

\begin{table}[!t]
    \centering    
    \caption{Accuracy, privacy leakage, and energy consumption performance compared to baselines: smaller $\text{FSIM}_{\text{total}}$ and $\overline{E}_{\text{total}}$ indicate better privacy protection and less energy consumption for environment setting A (See \reftab{diff-env-sp})} 
    \setlength{\tabcolsep}{1.5pt}
    \footnotesize
    \label{tab:baselineA}
    \vspace{0 in}
{
        \centering
        \begin{tabular}{llllll}
            \hline
            \textbf{Model} & \textbf{Dataset} & \textbf{System} & \textbf{Accuracy} & $\textbf{\text{FSIM}}_{\text{total}}$ & $\overline{\mathbf{E}}_{{\text{total}}}$ \textbf{(kJ)} \\\hline
            
             \multirow{12}{*}{\centering VGG16-BN}& \multirow{4}{*}{\centering CIFAR-10}  & \textbf{P3SL} & \textbf{0.90} \textbf{($\uparrow$)} & \textbf{2.52} \textbf{($\downarrow$)} & \textbf{4390} \textbf{($\downarrow$)}\\ 
                      &          & ASL        & 0.84                    & 2.55           & 6503           \\ 
                      &          & ARES        & 0.85                    & 2.57           & 6452           \\ 
                      &           & SSL         & 0.86                    & 2.63           & 8446           \\ \cline{2-6}
       
             & \multirow{4}{*}{\centering Fashion-MNIST}  & \textbf{P3SL} & \textbf{0.92} \textbf{($\uparrow$)} & \textbf{2.52} \textbf{($\downarrow$)}& \textbf{5953} \textbf{($\downarrow$)}\\ 
                      &                & ASL          & 0.85                    & 2.60           & 9443\\ 
                     &                & ARES          & 0.84                    & 2.58           & 9393\\ 
                      &                & SSL           & 0.84                    & 2.59           & 14562           \\ \cline{2-6}
            
                      & \multirow{4}{*}{\centering FLOWER-102}      & \textbf{P3SL} & \textbf{0.89} \textbf{($\uparrow$)}     & \textbf{2.52} \textbf{($\downarrow$)}& \textbf{617} \textbf{($\downarrow$)}\\ 
                      &                & ASL          & 0.84                    & 2.56           & 902           \\ 
                      &                & ARES          & 0.83                    & 2.54           & 894           \\ 
                      &                & SSL           & 0.86                    & 2.57           & 1335           \\ \hline
            
                     \multirow{12}{*}{\centering ResNet18} &  \multirow{4}{*}{\centering CIFAR-10}    & \textbf{P3SL} & \textbf{0.91} \textbf{($\uparrow$)} & \textbf{2.51} \textbf{($\downarrow$)}& \textbf{6980} \textbf{($\downarrow$)}\\ 
                      &                & ASL          & 0.86                    & 2.59           & 9896           \\ 
                      &                & ARES          & 0.85                    & 2.58           & 9994           \\ 
                      &                & SSL           & 0.84                    & 2.53           & 13582           \\ \cline{2-6}
            
             & \multirow{4}{*}{\centering Fashion-MNIST}    & \textbf{P3SL} & \textbf{0.92} \textbf{($\uparrow$)} & \textbf{2.51} \textbf{($\downarrow$)}& \textbf{9566} \textbf{($\downarrow$)}\\ 
                      &    & ASL          & 0.86                    & 2.59           & 13744           \\ 
                &     & ARES          & 0.88                    & 2.63           & 13588           \\ 
                      &                & SSL           & 0.85                    & 2.58           & 19942           \\ \cline{2-6}
            
                      & \multirow{4}{*}{\centering FLOWER-102}     & \textbf{P3SL} & \textbf{0.91} \textbf{($\uparrow$)} & \textbf{2.51} \textbf{($\downarrow$)} & \textbf{1004} \textbf{($\downarrow$)}\\ 
                      &               & ASL          & 0.84                    & 2.60           & 1482           \\ 
                      &               & ARES          & 0.84                    & 2.59           & 1421           \\ 
                      &                & SSL           & 0.88                    & 2.62           & 2005           \\ \hline

                      \multirow{12}{*}{\centering ResNet101} &\multirow{4}{*}{\centering CIFAR-10}        & \textbf{P3SL} & \textbf{0.92} \textbf{($\uparrow$)} & \textbf{2.50} \textbf{($\downarrow$)}& \textbf{7012} \textbf{($\downarrow$)}\\ 
                      &                & ASL          & 0.85                   & 2.59          & 10799           \\ 
                      &               & ARES          & 0.86                   & 2.61          & 11023           \\ 
                      &                & SSL           & 0.89                   & 2.61           & 12811           \\ \cline{2-6}
            
             & \multirow{4}{*}{\centering Fashion-MNIST}  & \textbf{P3SL} & \textbf{0.94} \textbf{($\uparrow$)} & \textbf{2.50} \textbf{($\downarrow$)}& \textbf{9467} \textbf{($\downarrow$)}\\ 
                      &       & ASL          & 0.87                    & 2.61           & 15077          \\ 
                      &      & ARES          & 0.87                    & 2.60           & 15231          \\ 
              
                      &                & SSL           & 0.91                    & 2.62           & 16870           \\ \cline{2-6}
            
                      &  \multirow{4}{*}{\centering FLOWER-102}    & \textbf{P3SL} & \textbf{0.91} \textbf{($\uparrow$)} & \textbf{2.45} \textbf{($\downarrow$)} & \textbf{1023} \textbf{($\downarrow$)}\\ 
                      &                & ASL          & 0.87                 & 2.59           & 1600           \\ 
                       &               & ARES          & 0.90                 & 2.63           & 1507           \\ 
                      &                & SSL           & 0.87                    & 2.60           & 2346           \\ \hline
        \end{tabular}
        \label{tab:baseline-env1}
    }
\end{table}

\subsection{RQ1: Privacy Leakage, Energy Consumption, and Global Model Accuracy}
We first evaluate the overall performance of \textsc{P3SL} in terms of privacy leakage, energy consumption, and global model accuracy compared to the baselines, under two heterogeneous environmental settings. 

\textbf{Privacy Leakage and Accuracy.} As shown in \reftab{baselineA}, \textsc{P3SL} consistently reduces overall privacy leakage while maintaining high global accuracy, leveraging sequential SL without overloading the server. Although all methods apply the same noise injection defense, \textsc{P3SL} outperforms baselines by jointly optimizing split points and privacy levels per client via bi-level optimization, ensuring a more appropriate trade-off between privacy protection and energy consumption. In contrast, the baselines do not account for privacy leakage when selecting split points, leading to inefficient protection. Some clients may be overprotected with large noise, especially when applied to deeper layers, which yields minimal privacy gains but significantly harms accuracy. \textsc{P3SL} also preserves personalized client models by avoiding parameter sharing, further enhancing privacy. By integrating privacy into personalized split point selection, \textsc{P3SL} enhances overall privacy protection without compromising much accuracy. While \textsc{P3SL} has a bit longer training time than PSL methods (e.g., 418s/epoch for ResNet18 with F-MNIST vs. 252s/epoch for ARES and ASL), it remains practical, and the improved accuracy from sequential SL justifies the additional cost.

\textbf{Energy Consumption.} As shown in \reftab{baseline}, \textsc{P3SL} reduces total energy consumption by up to 38.68\% and 59.12\% compared to ARES and SSL, respectively. ARES's high energy consumption results from its reliance on PSL, which requires devices to remain actively engaged in computation for extended periods while waiting for stragglers to finish training. Moreover, ARES lacks a parallel execution management strategy, further leading to energy inefficiency. In contrast, SSL's energy inefficiencies primarily stem from frequent communication overhead, as model parameters must be repeatedly transmitted to initialize adjacent clients' training. \textsc{P3SL} addresses these issues by employing a weighted aggregation technique, reducing the frequency of parameter uploads to the server and eliminating the need to distribute aggregated weights back to clients. Additionally, the use of sleep-wake scheduling minimizes energy consumption during idle time in the sequential training process for clients.

\begin{remark}
Personalized split points with personalized models in \textsc{P3SL} significantly reduce privacy leakage risks and system energy consumption while maintaining high accuracy.
\end{remark}

\begin{table}[!t]
    \centering    
    \caption{Accuracy, privacy leakage, and energy consumption performance for VGG16-BN with CIFAR-10 dataset compared to baselines for heterogeneous environment settings}
    \setlength{\tabcolsep}{5pt}
    \footnotesize
    \label{tab:baseline}

   {
        \centering
        \begin{tabular}{l|llll}
            \hline
            \textbf{Environment}  & \textbf{System} & \textbf{Accuracy} & $\textbf{\text{FSIM}}_{\text{total}}$ & $\overline{\mathbf{E}}_{{\text{total}}}$ \textbf{(kJ)} \\\hline
            
             \multirow{4}{*}{\centering Environment A} &   \textbf{P3SL} & \textbf{0.90} \textbf{($\uparrow$)} & \textbf{2.52} \textbf{($\downarrow$)} & \textbf{4390} \textbf{($\downarrow$)}\\ 
                                & ASL        & 0.84                    & 2.55           & 6503           \\ 
                                & ARES        & 0.85                    & 2.57           & 6452           \\ 
                                & SSL         & 0.86                    & 2.63           & 8446           \\ \hline

             \multirow{4}{*}{\centering Environment B} &   \textbf{P3SL} & \textbf{0.91} \textbf{($\uparrow$)} & \textbf{2.51} \textbf{($\downarrow$)} & \textbf{4433} \textbf{($\downarrow$)}\\ 
                               & ASL        & 0.86                    & 2.56           & 6299           \\ 
                                & ARES        & 0.87                    & 2.57           & 6168           \\ 
                                 & SSL         & 0.89                    & 2.60           & 8126           \\ \hline

        \end{tabular}
    }    

\end{table}
\subsection{RQ2: Impact of Heterogeneous Environment Settings} 
To evaluate the impact of environmental settings, we compare the total system privacy leakage $\text{FSIM}_{\text{total}}$ and accuracy performance of P3SL, ARES, and SSL systems using three models and three datasets under two heterogeneous environmental conditions, as shown in \reftab{baseline}.
%\bo{you may want to switch this subsection with Subsection VI.C (similarly, RQ2 vs. RQ3) as what you discuss here is based on \reftab{baseline}} \Comment{Revised} 
While different environmental settings result in varied energy profiles and optimal split points for each client, \textsc{P3SL} achieves the highest accuracy and lowest total system privacy leakage compared to baselines. This demonstrates the effectiveness of its bi-level optimization design in maintaining high model accuracy and minimizing system privacy leakage across heterogeneous environments.

\begin{remark}
P3SL demonstrates robustness in maintaining both high accuracy and low privacy leakage across varying environments through its adaptive noise adjustment and bi-level optimization mechanism.
\end{remark}

\begin{figure}[!t]
    \centering
    \subfigure[Environment setting A]{\includegraphics[width = 0.415\columnwidth]{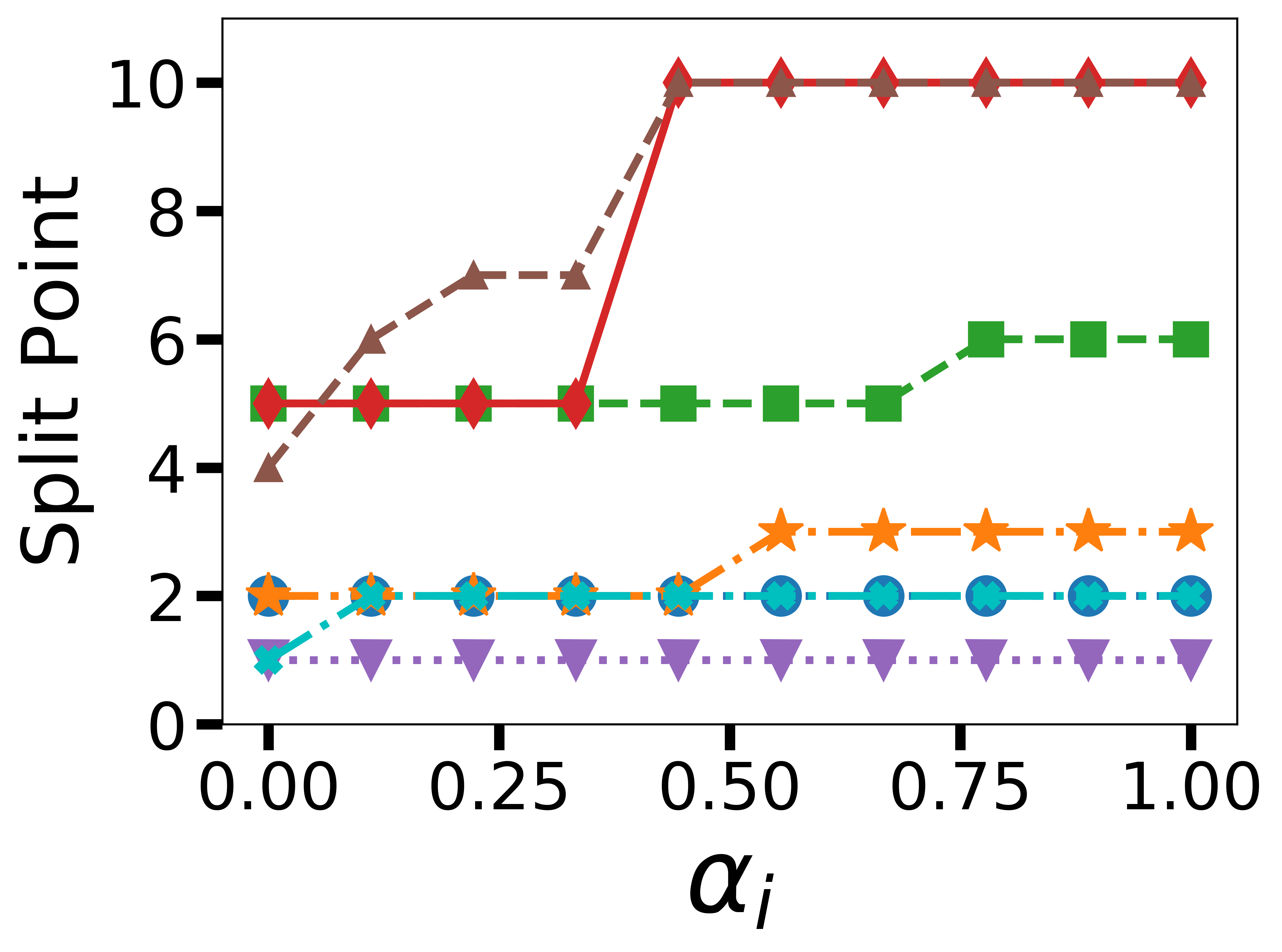}
    \label{fig:optimal-sl-env1}}\hspace{-2mm}
    \subfigure[Environment setting B]{\includegraphics[width = 0.56\columnwidth]{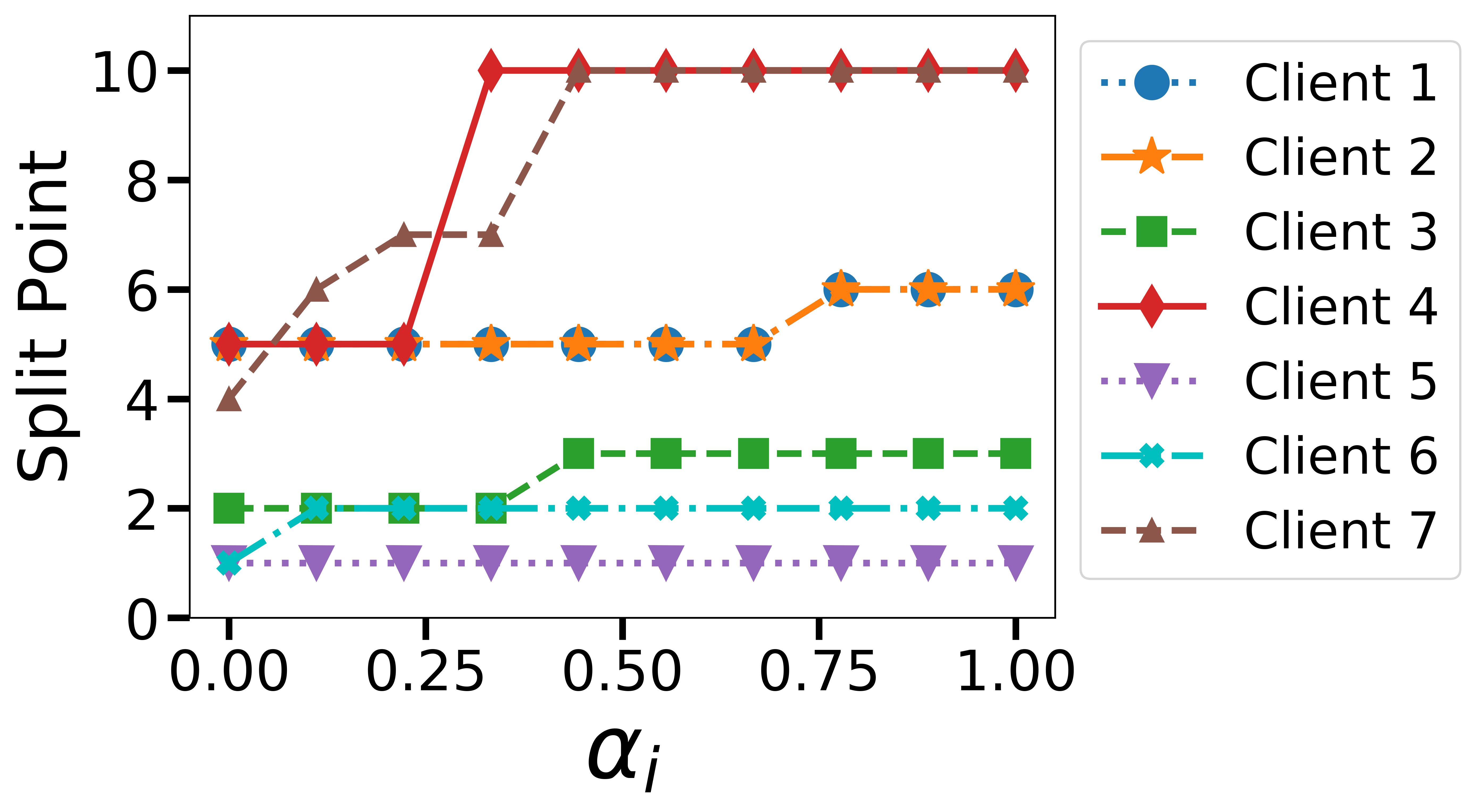}
    \label{fig:optimal-s1-env2}}
    \vspace{-1mm}
    \caption{Optimal split point selection for different personalized privacy sensitivity coefficient $\alpha_i$}
    \vspace{-4mm}
    \label{fig:optimal-sl}
\end{figure}

\subsection{RQ3: Personalized Privacy Evaluation}
We evaluate how \textsc{P3SL} enables personalized privacy protection for clients with heterogeneous resource constraints and privacy requirements. We further analyze the impact of varying the privacy sensitivity coefficient \(\alpha_i\). 

\textbf{Personalized Noise Injection and Privacy Requirements.} As shown in Table~\ref{tab:personalized-privacy-p3sl}, each client in \textsc{P3SL} selects its split point based on computational resources, privacy preference, and power constraints. Personalized privacy protection in \textsc{P3SL} encompasses two components: personalized noise injection and the personalized privacy requirement, represented by the privacy sensitivity coefficient $\alpha_i$. Clients with higher \(\alpha_i\) prioritize privacy protection by selecting deeper split points, which inherently offer stronger privacy due to more number of layers involved. For instance, Client 4 and Client 7, with high $\alpha_i$, select split point 10, requiring minimal noise injection (\(\sigma = 0.02\)) to achieve privacy preservation. Their corresponding FSIM reduces from \(0.366 \rightarrow 0.355\) and \(0.366 \rightarrow 0.354\), respectively. In contrast, clients with lower \(\alpha_i\) prioritize energy efficiency, selecting shallower split points where privacy leakage risks are higher. To compensate, these clients require higher levels of noise injection to achieve similar FSIM reductions. For example, Client 2 and Client 6, with low $\alpha_i$, select split point 3 (\(\sigma = 1.15\)) and split point 2 (\(\sigma = 1.65\)), reducing FSIM from \(0.428 \rightarrow 0.356\) and \(0.532 \rightarrow 0.360\), respectively. This approach reflects personalized privacy protection, as clients adapt their split point selection and noise injection based on their personalized privacy preference (i.e., $\alpha_i$), significantly reducing FSIM scores across all clients. 

\textbf{Impact of Personalized Privacy Sensitivity Coefficient \(\alpha_i\).}  
We further analyze the impact of the privacy sensitivity coefficient \(\alpha_i\) by varying it from \(0 \rightarrow 1\) in intervals of 0.1 for each client. As shown in Fig.~\ref{fig:optimal-sl} 
% \bo{need to switch the location of Fig.~6 and Fig.~7} \Comment{Changed}
under two different environment settings (\reftab{diff-env-sp}), clients with smaller ($\alpha_i$) values tend to prioritize energy efficiency, resulting in the selection of shallower split points. Conversely, as \(\alpha_i\) increases, clients prefer deeper split points to enhance privacy protection. Additionally, some split points in the middle appear to be sweet spots, offering smaller intermediate representations that save communication energy and provide better privacy protection compared to shallower split points. 

\begin{remark}
\textsc{P3SL} 
suggests that personalized privacy preservation can significantly mitigate privacy risks tailored to each heterogeneous resource-constrained client and environments.
\end{remark}

\begin{table}[!t]
    \centering

    \footnotesize
    \caption{The effect of heterogeneous split points and personalized privacy protection using VGG16-BN in environment setting A. FSIM represents the privacy leakage level, where higher FSIM indicates greater privacy leakage. Notably, a small difference in FSIM scores highlights significant variations in privacy leakage levels. $\alpha_i$ represents the personalized privacy sensitivity coefficient. The FSIM values are shown both before and after noise injection.}
    % \vspace{1mm}
    \label{tab:personalized-privacy}    
    {
        \setlength{\tabcolsep}{1pt}
        \centering
        \begin{tabular}{ccccc}
            \hline
            \textbf{Client ID} & \textbf{Split Point} & $\alpha_i$ & \textbf{Noise Injection} & \textbf{FSIM (Before $\rightarrow$ After)} \\
            \hline
            1 (Jetson Nano) & 2 & 0.4 & $Lap(0, 1.65^2)$ & 0.53 $\rightarrow$ \textbf{0.36} \\
            2 (Jetson Nano) & 3 & 0.2 & $Lap(0, 1.15^2)$ & 0.43 $\rightarrow$ \textbf{0.36} \\
            3 (Jetson Nano) & 5 & 0.5 & $Lap(0, 0.2^2)$ & 0.41 $\rightarrow$ \textbf{0.37} \\
            4 (Jetson Nano) & 10 & 0.9 & $Lap(0, 0.02^2)$ & 0.37 $\rightarrow$ \textbf{0.36} \\
            5 (Raspberry Pi) & 1 & 0.7 & $Lap(0, 2.15^2)$ & 0.53 $\rightarrow$ \textbf{0.37} \\
            6 (Raspberry Pi) & 2 & 0.3 & $Lap(0, 1.65^2)$ & 0.53 $\rightarrow$ \textbf{0.36} \\
            7 (Laptop) & 10 & 0.8 & $Lap(0, 0.02^2)$ & 0.37 $\rightarrow$ \textbf{0.35} \\
            \hline
        \end{tabular}%
        \label{tab:personalized-privacy-p3sl}
    }

\vspace{-3mm}
\end{table}

\begin{figure}[!t]
    \centering
    \subfigure[Schedule of clients attendance]{\includegraphics[width = 0.5\columnwidth]{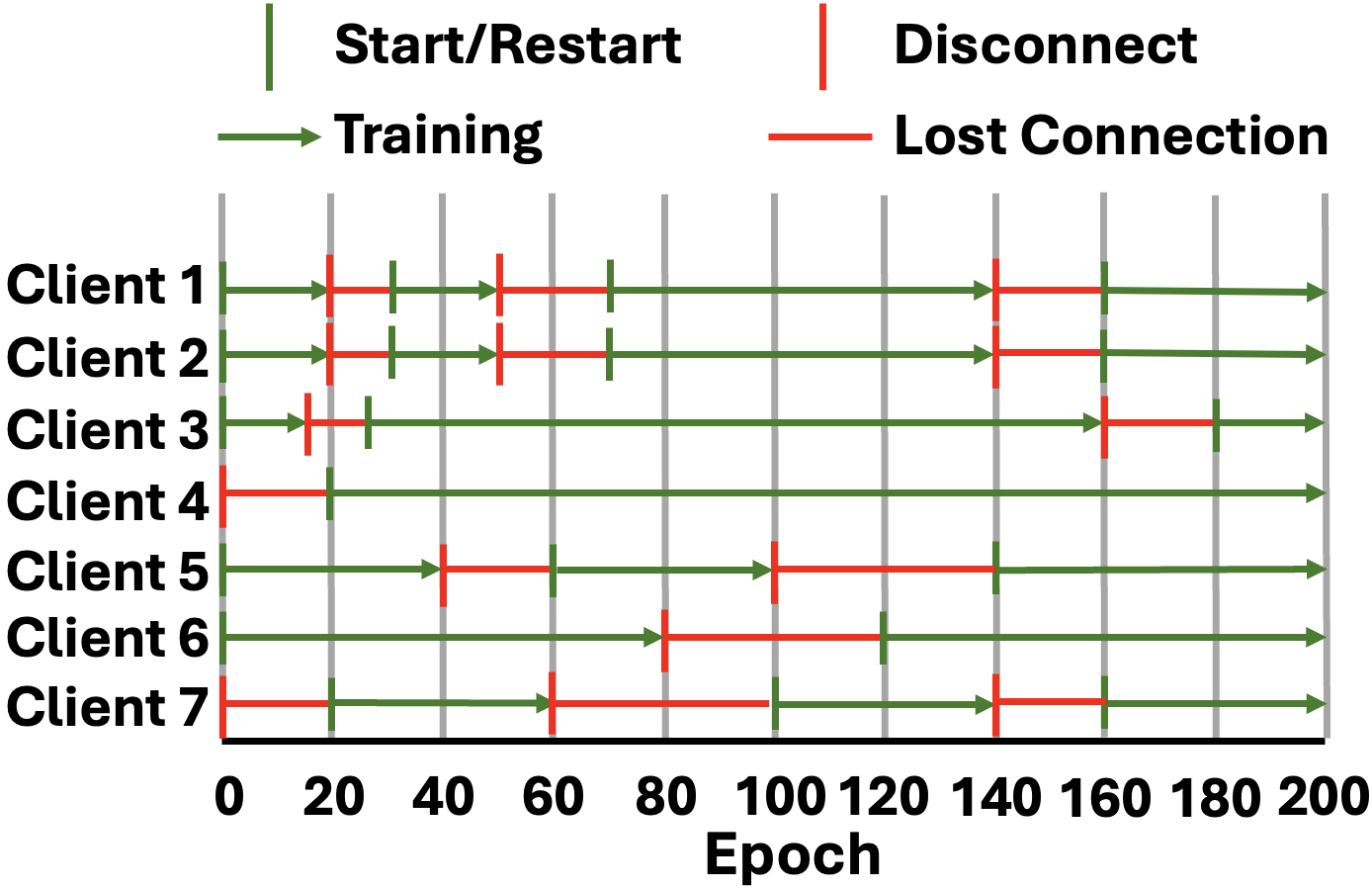}
    \label{fig:adapt-schedule}}\hspace{-1mm}
    \subfigure[Accuracy dynamics %across datasets 
    under varying network conditions]{\includegraphics[width = 0.47\columnwidth]{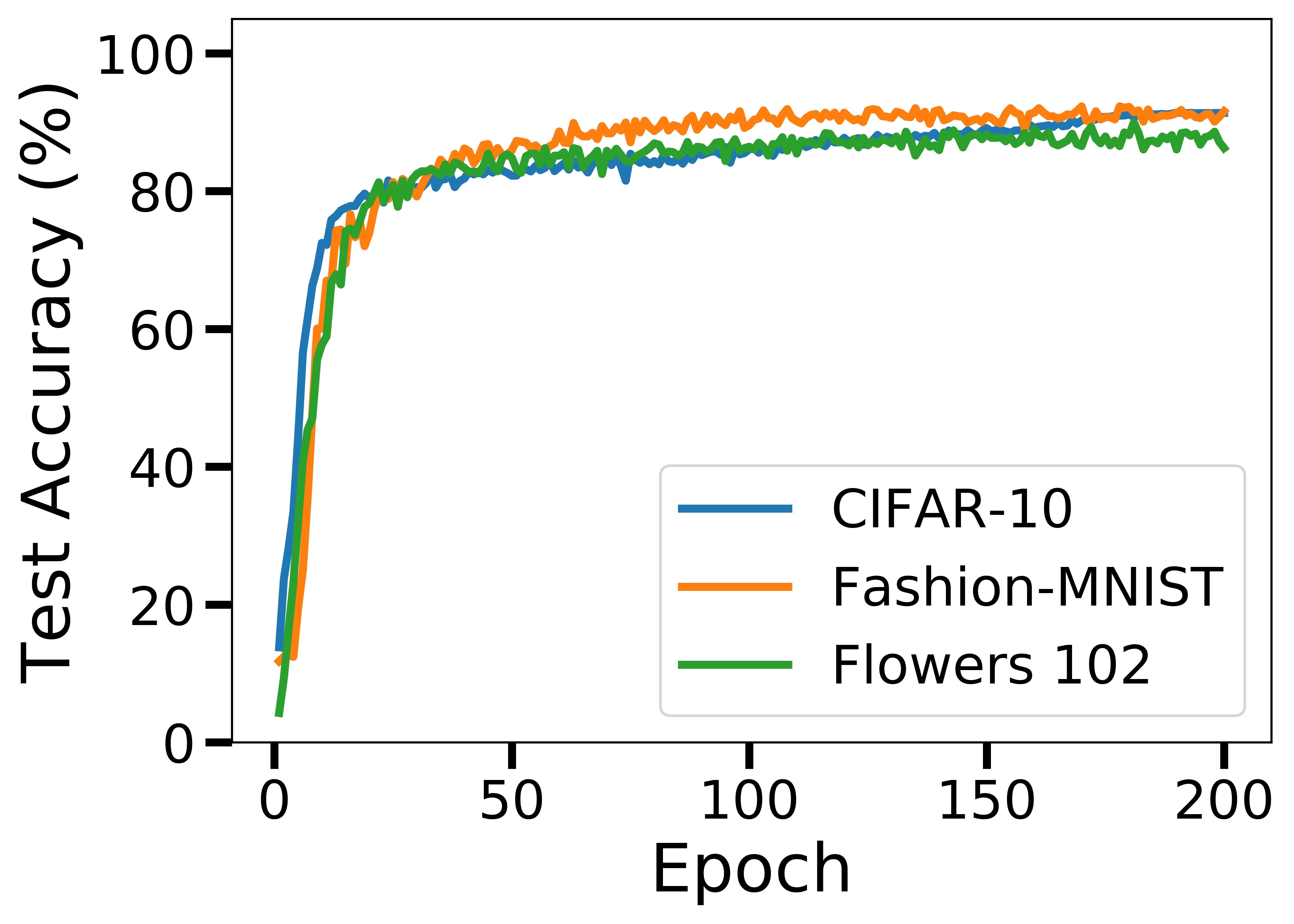}
    \label{fig:varying-network}}
    \vspace{-1mm}
    \caption{Adaptability to Varying Network Connections}
    \vspace{-4mm}
    \label{fig:adaptability}
\end{figure}

\Comment{\subsection{RQ4: Adaptability under Network Fluctuations} 
We examine the adaptability of the \textsc{P3SL} system to varying network connection conditions among client devices, where some clients may lose connection during certain training periods, while some new clients may join at specific epochs. In such cases, the system skips the training process with disconnected clients and proceeds with the available ones. We conduct the experiment under the attendance schedule of four clients, as shown in~\reffig{adapt-schedule}, jointly training the VGG16-BN model with environment setting B using three separate datasets. For instance, at epoch 20, Clients 5 and 6 continue training, while Clients 1, 2, and 3 lose connection, and Clients 4 and 7 newly join the training process. %\Wei{Changed here to the end of paragraph:} 
\reffig{varying-network} presents the evolution of global model accuracy across three datasets. Although convergence is slower compared to the static full-client scenario, \textsc{P3SL} ultimately reaches the same performance level. This demonstrates the framework’s robustness to dynamic client participation and highlights that sequential training effectively accommodates new clients without disrupting the overall training trajectory.}

\Comment{\begin{remark}
The adaptability of \textsc{P3SL} to varying network conditions among client devices highlights its resilience to client disconnections and dynamic client additions.
\end{remark}}

\subsection{RQ5: System Robustness in Large-Scale Setting} 
\Comment{We evaluate the scalability and robustness of the \textsc{P3SL} system under increasing numbers of participating clients. Specifically, we simulate training with 5, 10, 15, and 20 clients using the VGG16-BN, ResNet18, and ResNet101 models on the CIFAR-10 dataset. Each client follows the same local training configuration, while employs different split points and privacy protection levels. As shown in \reftab{scalability}, \textsc{P3SL} maintains high global model accuracy across all scales, indicating strong scalability in heterogeneous environments. However, we observe a slight decline in accuracy and an increase in system-wide privacy leakage as the number of clients grows. This degradation results from the fixed overall privacy budget being divided among more clients, which reduces the amount of noise each client can inject without exceeding the system’s accuracy threshold. Consequently, the per-client privacy protection weakens, and the system becomes more susceptible to potential data reconstruction attacks.}

\Comment{\begin{remark} \textsc{P3SL} demonstrates strong scalability and robustness as the number of clients increases, though a modest decline in accuracy and privacy protection is observed due to the fixed privacy budget being shared across more participants. \end{remark}}

\begin{table}[!t]
    \centering    
    \caption{Accuracy and privacy leakage performance of three models on the CIFAR-10 dataset under simulation with 5, 10, 15, and 20 devices}
    \setlength{\tabcolsep}{5pt}
    \footnotesize
    \label{tab:scalability}

   {
        \centering
        \begin{tabular}{cccc}
            \hline
            \textbf{Model}  & \textbf{\# of Devices} & \textbf{Accuracy} & $\textbf{\text{FSIM}}_{\text{total}}$  \\\hline
            
             \multirow{4}{*}{\centering VGG16-BN} &   5 & 0.92 & 2.50\\ 
                                & 10        & 0.92                    & 2.54                    \\ 
                                & 15        & 0.90                    & 2.59                    \\ 
                                & 20         & 0.90                    & 2.63                 \\ \hline

             \multirow{4}{*}{\centering ResNet18} &   5 & 0.91 & 2.48 \\ 
                               & 10        & 0.90                    & 2.51                   \\ 
                                & 15        & 0.89                    & 2.57                    \\ 
                                 & 20         & 0.89                    & 2.61                      \\ \hline
       
             \multirow{4}{*}{\centering ResNet101} &   5 & 0.92 & 2.50 \\ 
                               & 10        & 0.91                    & 2.51                   \\ 
                                & 15        & 0.90                    & 2.57                    \\ 
                                 & 20         & 0.90                    & 2.61                      \\ \hline            

        \end{tabular}
    }    

\end{table}

\subsection{RQ6: Impact of Membership Inference Attack (MIA)}
\Comment{In split learning, an adversary typically attempts to perform a Membership Inference Attack (MIA) by determining whether a specific data point was part of a client’s training dataset~\cite{MIA}. This is commonly done by training a shadow model that replicates the target client’s training process and then using the shadow model's behavior to train an attack classifier. To evaluate the robustness of \textsc{P3SL} under such threats, we simulate MIAs across various conditions using the VGG16-BN model on CIFAR-10. First, we assess MIA accuracy across 10 split points when the shadow and target models are trained at the same stage (epoch 50), without L2 regularization. As shown in~\reftab{three_mia_subtables}(a), attack accuracy consistently exceeds 74\%, with peaks over 77\%, indicating a high MIA risk in this setting.}

\Comment{Next, we investigate the effect of training-stage mismatch between the shadow model and the target model in~\reftab{three_mia_subtables}(b). When the shadow model is trained at a different stage from the target client model (i,e., the different number of epochs), the attack accuracy drops to around 50\%, even without applying L2 regularization. This suggests that training misalignment itself weakens the attacker’s ability to infer membership information. Meanwhile, the results show that MIA is most effective when both models are trained at the same stages—achieving attack accuracy as high as 77.2\%. To mitigate the attack, we apply L2 regularization with $\lambda=0.08$ during split learning. As shown in~\reftab{three_mia_subtables}(c), L2 regularization significantly reduces MIA accuracy to approximately 50\% across all aligned stages. This indicates that L2 regularization limits overfitting and reduces leakage in the intermediate representations. Together, these findings demonstrate that \textsc{P3SL} is resilient to MIAs.}
 
\Comment{\begin{remark}
\textsc{P3SL} is robust against membership inference attacks, achieving random-guess-level attack performance when leveraging L2 regularization.
\end{remark}}

%%--------table--------
\begin{table}[t]
\centering
\caption{
MIA results on VGG16-BN with CIFAR-10, simulated on a server coordinating 7 clients with evenly distributed data and different split points. 
(a) shows MIA accuracy across 10 split points when the training stages of shadow model and target model are aligned;
(b) examines the impact of training stage misalignment at split point 5;
(c) shows MIA accuracy after applying L2 regularization ($\lambda = 0.08$) when training stages are aligned.}

\label{tab:three_mia_subtables}

% -------- Subtable (a) --------
\vspace{0.5em}
(a) MIA Accuracy for 10 Split Points shadow and target models are trained at the same stage (epoch 50) w/o L2 regularization
\vspace{0.3em}

\begin{tabular}{c c}
\toprule
\textbf{Split Point} & \textbf{MIA Accuracy} \\
\midrule
1  & 0.77 \\
2  & 0.75 \\
3  & 0.75 \\
4  & 0.75 \\
5  & 0.76 \\
6  & 0.76 \\
7  & 0.75 \\
8  & 0.76 \\
9  & 0.75 \\
10 & 0.76 \\
\bottomrule
\end{tabular}

\vspace{1.2em}

% -------- Subtable (b) --------
(b) MIA accuracy under varying training stage alignment w/o L2 regularization (Bold entries indicate cases with effective MIA)
\vspace{0.3em}

\begin{tabular}{cc c c}
\toprule
\multicolumn{2}{c}{\textbf{Trained Stage (Epoch)}} & \multirow{2}{*}{\textbf{MIA Accuracy}} \\
\cmidrule(lr){1-2}
\textbf{Shadow Model} & \textbf{Target Model} & \\
\midrule
\textbf{30} & \textbf{30} & \textbf{0.77 ($\uparrow$)} \\
30 & 50 & 48.9 \\
30 & 70 & 51.0 \\
50 & 30 & 50.1 \\
\textbf{50} & \textbf{50} & \textbf{0.76 ($\uparrow$)} \\
50 & 70 & 49.2 \\
70 & 30 & 50.9 \\
70 & 50 & 48.2 \\
\textbf{70} & \textbf{70} & \textbf{0.76 ($\uparrow$)} \\
\bottomrule
\end{tabular}

\vspace{1.2em}

% -------- Subtable (c) --------
(c) MIA Accuracy with L2 Regularization ($\lambda = 0.08$) where the shadow and target models are trained at the same stage
\vspace{0.3em}

\begin{tabular}{cc c}
\toprule
\multicolumn{2}{c}{\textbf{Trained Stage (Epoch)}} & \multirow{2}{*}{\textbf{MIA Accuracy}} \\
\cmidrule(lr){1-2}
\textbf{Shadow Model} & \textbf{Target Model} & \\
\midrule
30 & 30 & 0.77 $\rightarrow$ \textbf{0.50} \\
50 & 50 & 0.76 $\rightarrow$ \textbf{0.51} \\
70 & 70 & 0.76 $\rightarrow$ \textbf{0.50} \\
\bottomrule
\end{tabular}
\end{table}

\begin{figure}[!t]
    \centering
    \includegraphics[width=0.7\linewidth]{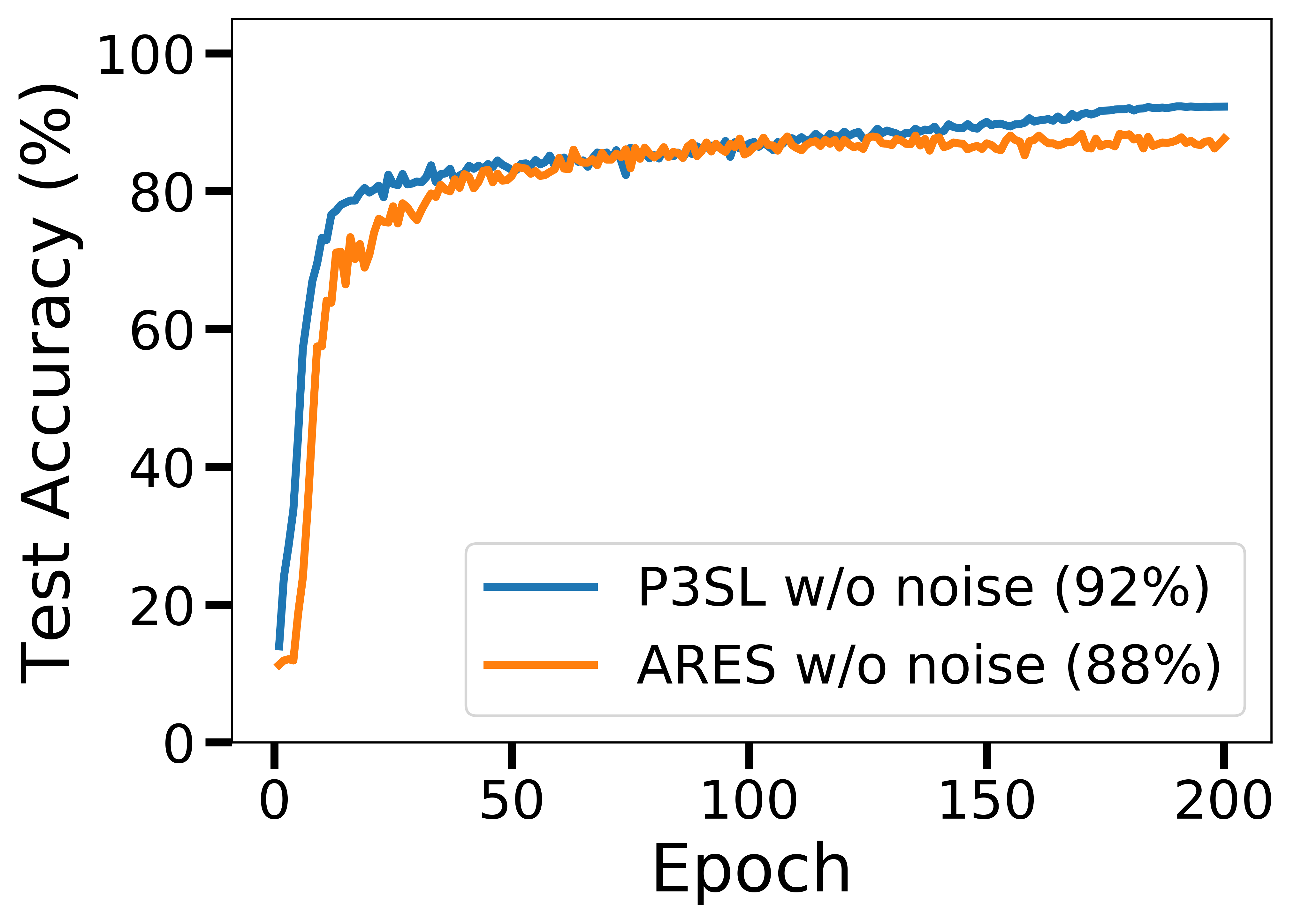}
    \vspace{-1mm}
    \caption{Accuracy comparison between P3SL and ARES without privacy protection, where the final accuracy is indicated in the legend}
    \label{fig:baseline-comparison}
    \vspace{-4mm}
\end{figure}

%%----------ablation studies----------------
\Comment{\subsection{Ablation Studies} 
% \Wei{I changed this section entirely}
\textbf{Effectiveness of \textsc{P3SL} Sequential Training Architecture.} To validate the effectiveness of the proposed \textsc{P3SL} sequential training architecture, we compare it against \textsc{ARES}, a parallel split learning baseline. This ablation study is designed to isolate the contribution of our sequential training strategy in heterogeneous environments with personalized privacy preservation and to evaluate \textsc{P3SL}'s impact on both accuracy and training efficiency. For a fair comparison, neither \textsc{P3SL} nor \textsc{ARES} applies privacy protection to intermediate representations during training. Both support heterogeneous split points across clients, and we adopt the same optimal split points and the customized VGG-16 model proposed in \textsc{ARES} for five clients on CIFAR-10. As shown in \reffig{baseline-comparison}, \textsc{P3SL} achieves a 4\% improvement in accuracy over ARES. Unlike \textsc{ARES}’s parallel training design with synchronous aggregation at every epoch, \textsc{P3SL} employs a server-centric sequential training paradigm with periodic aggregation every five epochs. Importantly, \textsc{P3SL} avoids inter-client model transmission, which is required in prior SSL approaches. In SSL, each client must inherit and continue training from the previous client’s model, and the system assumes uniform split points across clients. In contrast, \textsc{P3SL} allows clients to train independently with personalized split points and shares model updates only with the central server. This design reduces communication overhead and enables more flexible participation, while still allowing the shared server to learn from diverse representations accumulated over multiple rounds. The observed improvement in accuracy and reduced aggregation frequency highlight the effectiveness of the \textsc{P3SL} training architecture.}

\Comment{\begin{remark}
The proposed \textsc{P3SL} sequential training architecture outperforms the parallel baseline \textsc{ARES}, achieving higher accuracy and lower aggregation frequency while supporting personalized split points without inter-client model sharing.
\end{remark}}

\section{Discussion} \label{sec:discussion}
In this section, we discuss additional privacy concerns and server profiling costs associated with \textsc{P3SL}.

\textbf{Additional Privacy Concerns.} 
While \textsc{P3SL} ensures robust privacy protection, additional privacy considerations arise from model sharing between clients and the server during the aggregation process, as well as label sharing during training. Model sharing exposes the system to specific threat models, such as Gradient Leakage Attack (GLA)~\cite{FL_GLA}. However, in the threat model considered in this work, label leakage does not pose any impact.

\Comment{\textit{Gradient Leakage Attack (GLA).}
GLA exploits gradient updates derived from the shared model between clients and the server during the aggregation process to reconstruct the original input data. In the \textsc{P3SL} design, noise injection is applied to the intermediate outputs sent between clients and the server, significantly reducing the adversary's ability to reconstruct data effectively~\cite{Gradient_leakage}.}

\Comment{\textit{Label Leakage.}
Label leakage occurs when label information is shared between clients and the server during training, potentially exposing sensitive information. However, the threat model considered in this work is not affected by label leakage. Existing solutions, such as U-shaped SL~\cite{ushape1}, allow clients to retain label locally without sharing it with the server. While \textsc{P3SL} does not specifically address label leakage, such techniques can be integrated into the framework in future extensions, further enhancing privacy protection.}

\textbf{Server Profiling Costs.} \textsc{P3SL} introduces slight computational overhead on the server for profiling the \textit{Privacy Leakage Table}. However, this process only performs once at the start and relies on a single dataset, as privacy leakage profiling is model-dependent rather than dataset-specific. By normalizing privacy leakage values, \textsc{P3SL} ensures consistent and fair privacy protection across diverse datasets. Centralizing this profiling task on the server further reduces the need for individual client profiling, conserving energy and computational resources on client devices. This step is critical for maintaining fair and consistent privacy protection for all clients while optimizing resource efficiency.

\Comment{\textbf{Client Selection.} Some methods~\cite{Huang_AL} allow the server to choose clients with qualified data to improve model training performance in distributed learning. This could be considered for SL to enhance global model robustness and performance. However, ensuring privacy fairness with non-IID data is challenging since each client may have data of varying importance. This is an interesting topic for future exploration, where actively selecting clients with higher quality data for \textsc{P3SL} could further improve the system's training performance.}

\section{Conclusion}

We proposed P3SL, a novel SL framework that allows clients to maintain personalized privacy protection, tailored to their heterogeneous resource constraints and deployment environments. The P3SL framework incorporated a sequential training architecture coupled with a weighted aggregation technique, allowing each edge device to maintain personalized local models, personalized privacy protection and heterogeneous split points. Additionally, the P3SL conducted bi-level optimization modeling to allow each client to strategically select optimal personalized split points, effectively balancing energy efficiency, privacy leakage risk, and consistently maintaining high accuracy. We implemented and deployed P3SL on a variety of edge devices, including Jetson Nano, Raspberry Pi, and laptops, and conducted extensive experimental evaluations. The evaluation results consistently demonstrated that P3SL can effectively achieve better personalized privacy protection and reduced energy consumption while consistently sustaining high accuracy, proving its efficacy in heterogeneous, resource-constrained distributed learning environments. 

For future work, P3SL's applicability could be extended beyond image classification models to include transformer architectures and other data types, such as tabular data, time series, and more. Additionally, incorporating advanced client selection mechanisms, which prioritize clients with high-quality data, presents an opportunity to further enhance global model robustness and training performance.

\bibliographystyle{IEEEtran}
\bibliography{bibliography}

\ifCLASSOPTIONcaptionsoff
  \newpage
\fi

\end{document}